%% file: main.tex
\documentclass[10pt,twocolumn,letterpaper]{article}

\usepackage{cvpr}              

\input{preamble}

\definecolor{cvprblue}{rgb}{0.21,0.49,0.74}
\usepackage[pagebackref,breaklinks,colorlinks,citecolor=cvprblue]{hyperref}

\def\paperID{46} 
\def\confName{}
\def\confYear{}

\title{\method: \\ Generating Consistent Animated Characters using Image Diffusion Models}

\author{Abdelrahman Eldesokey\\
KAUST, Saudi Arabia\\
{\tt\small abdelrahman.eldesokey@kaust.edu.sa}
\and
Peter Wonka\\
KAUST, Saudi Arabia\\
{\tt\small peter.wonka@kaust.edu.sa}
}

\begin{document}

\twocolumn[{%
\renewcommand\twocolumn[1][]{#1}%
\maketitle
\begin{center}
    \centering
    \captionsetup{type=figure}
    \includegraphics[width=\textwidth]{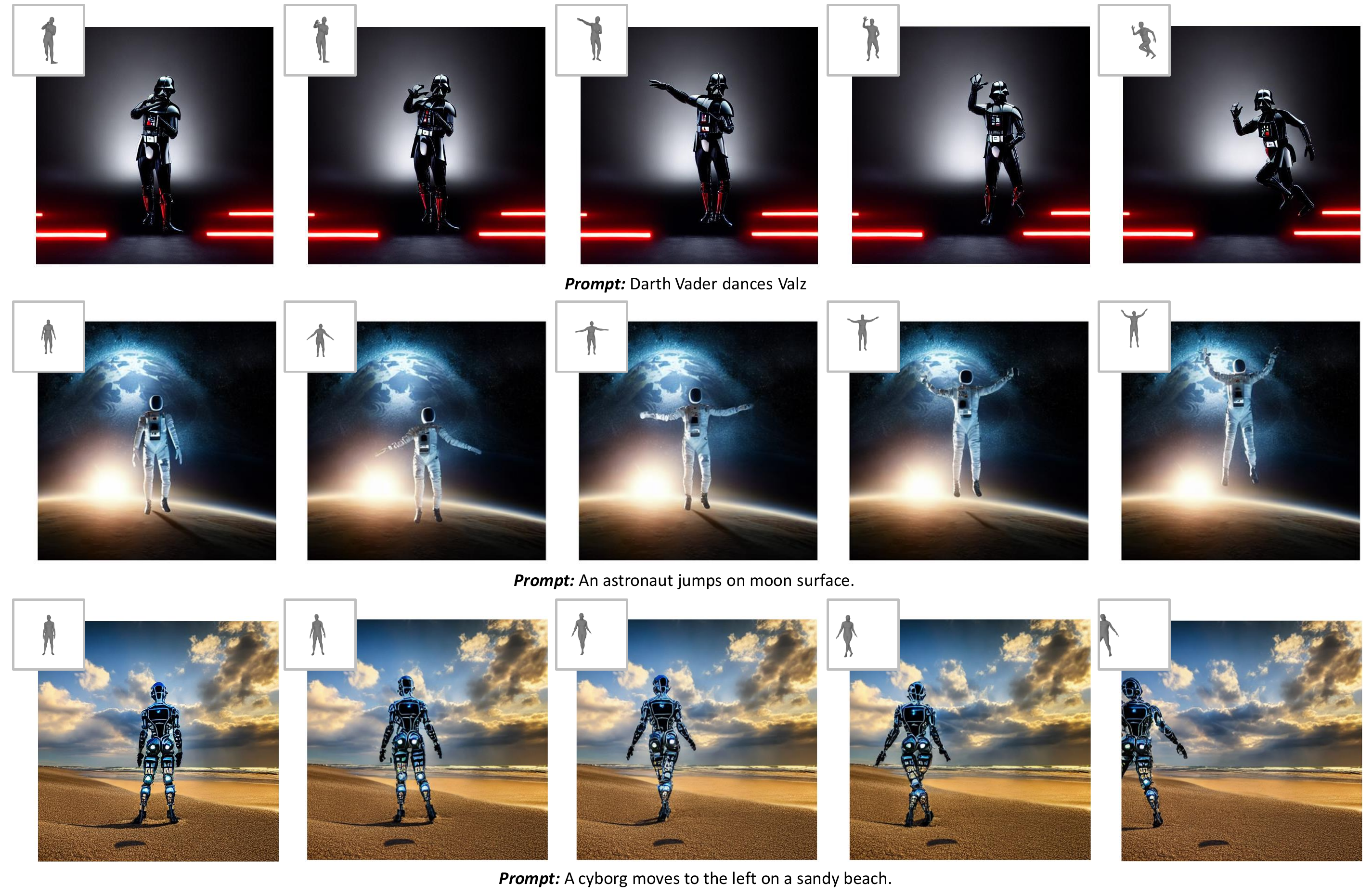}
    \captionof{figure}{\method produces temporally consistent videos of animated characters using pre-trained Motion and Text-to-Image (T2I) diffusion models given only a textual prompt.}
    \label{fig:teaser}
\end{center}%
}]

\input{sec/0_abstract}    
\input{sec/1_intro}

\input{sec/2_related}

\input{sec/3_method}

\input{sec/4_exp}
\input{sec/5_conclusion}

{
    \small
    \bibliographystyle{ieeenat_fullname}
    \bibliography{zbib}
}

\input{X_suppl}

\end{document}

%% file: preamble.tex
\usepackage[dvipsnames]{xcolor}
\usepackage{array}
\usepackage{algorithm}
\usepackage{algpseudocode}
\usepackage{multirow}

\newcommand{\alphatm}{\alpha_{t-1}}
\newcommand{\xzh}{\hat{x}_{0}^{i,t}}
\newcommand{\prompt}[1]{{\small \emph{Prompt}: ``#1"}}
\newcommand{\method}{\textsc{LatentMan }}

\newcolumntype{L}[1]{>{\raggedright\let\newline\\\arraybackslash\hspace{0pt}}m{#1}}
\newcolumntype{C}[1]{>{\centering\let\newline\\\arraybackslash\hspace{0pt}}m{#1}}
\newcolumntype{R}[1]{>{\raggedleft\let\newline\\\arraybackslash\hspace{0pt}}m{#1}}


%% file: sec/0_abstract.tex
\begin{abstract}
We propose a zero-shot approach for generating consistent videos of animated characters based on Text-to-Image (T2I) diffusion models.
Existing Text-to-Video (T2V) methods are expensive to train and require large-scale video datasets to produce diverse characters and motions. 
At the same time, their zero-shot alternatives fail to produce temporally consistent videos with continuous motion.
We strive to bridge this gap, and we introduce \method that leverages existing text-based motion diffusion models to generate diverse continuous motions to guide the T2I model.
To boost the temporal consistency, we introduce the Spatial Latent Alignment module that exploits cross-frame dense correspondences that we compute to align the latents of the video frames.
Furthermore, we propose Pixel-Wise Guidance to steer the diffusion process in a direction that minimizes visual discrepancies between frames.
Our proposed approach outperforms existing zero-shot T2V approaches in generating videos of animated characters in terms of pixel-wise consistency and user preference.
Project page \url{https://abdo-eldesokey.github.io/latentman/}.
\end{abstract}

\begin{figure*}[!ht]
  \centering
  \includegraphics[width=\textwidth]{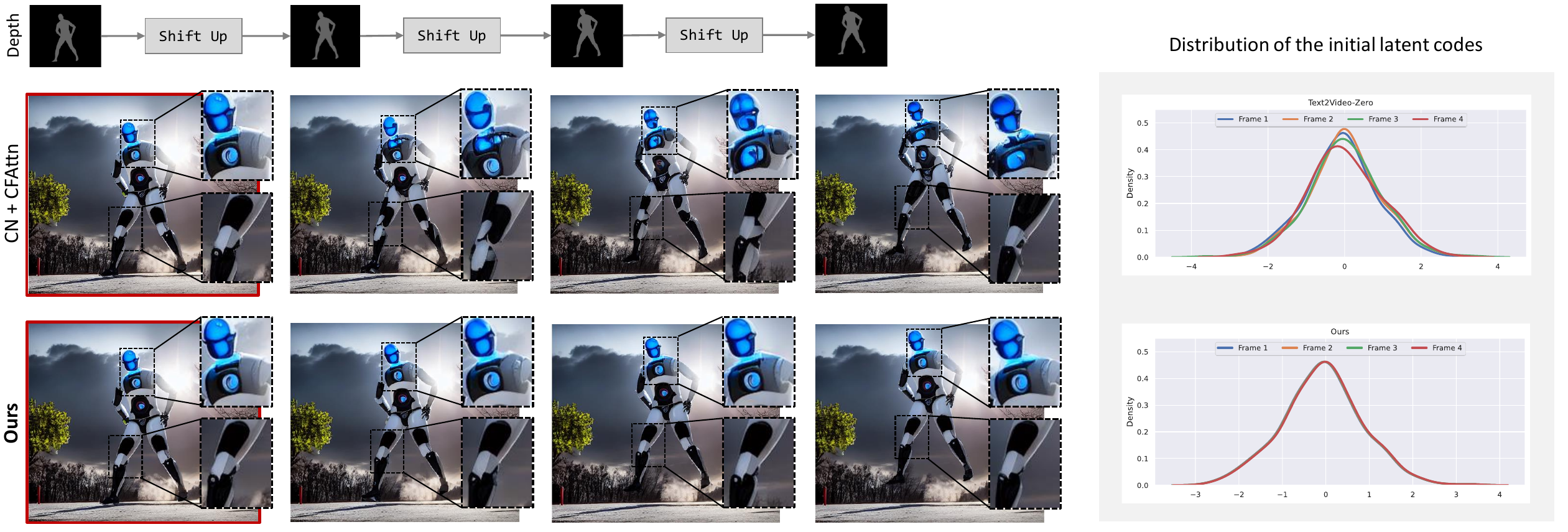}
  \caption{\emph{Cross-Frame Attention} (CFAttn) is adopted by multiple zero-shot T2V approaches to generate globally consistent video frames. However, when the conditioning signal (the depth map) changes, \eg shifted up, the fine details (shown in the insets) tend to vary between frames. We find that this is caused by the distributional shift of the initial latent codes that are aligned with the character, as shown on the plot to the right. Our proposed approach attempts to align the latent codes in a zero-shot manner, eliminating the distribution shift and producing consistent images. *CN refers to ControlNet }
  \label{fig:intro}
\end{figure*}

%% file: sec/1_intro.tex
\section{Introduction}
\label{sec:intro}


Generating visual assets of human characters is a prominent task in the realm of image and video synthesis, with many applications in movie production, art, and fashion.
This task aims to generate high-quality and diverse images/videos of human characters that adhere to some given conditions, \eg textual prompts and human poses.
Text-to-Image (T2I) diffusion models \cite{sd,dalle2,imagen} revolutionized this endeavor as they can generate high-quality images of human characters conditioned on user-provided textual prompts. 
ControlNet \cite{controlnet} allowed further control over the generated images through various conditioning signals such as depth maps, human poses, and edge maps.

For generating videos of human characters, Text-to-Video (T2V) diffusion models \cite{videoldm,blattmann2023stable,imagen_video,makeavideo} are evolving rapidly, but there are several complexities associated with them.
For instance, learning the motion dynamics (\eg the human body), finding sufficiently large datasets, and fulfilling their excessive computational needs.
As an example, the largest publicly available video dataset encompasses only 10 million videos \cite{webvid}, and it requires up to 48 A100-80GB GPUs to train VideoLDM \cite{videoldm} on this dataset.
Therefore, a growing direction of research attempts to democratize this task by leveraging T2I models to generate videos in \emph{few-} to \emph{zero-shot} manner.

One category of approaches \cite{tokenflow2023,yang2023rerender,fatezero,tuneavideo} adopts a Video-to-Video (V2V) scheme that relies on a reference video to generate a target video with modified contents.
However, these approaches require the user to provide the reference video, which can be difficult and inconvenient to find.
Alternatively, Text2Video-Zero \cite{tuneavideo} proposed to generate videos based only on a textual prompt where the motion dynamic is simulated by applying translation vectors to the latent codes of the first frame.
The temporal consistency was achieved by converting the \emph{self-attention} modules of the T2I UNet, which encodes the visual style, to \emph{cross-frame} attention. 
This enforces the T2I model to generate video frames that are visually consistent.
Nonetheless, the generated videos lack any motion continuity and only show random variations of the same object.
Moreover, a closer look at the generated frames shows that the temporal consistency is rather global, and fine details tend to change.

To illustrate this observation, we conduct a controlled experiment where we render a SMPL human model \cite{smpl} to obtain a depth map of a human.
We use this depth map and a textual prompt as conditions for ControlNet to generate a \emph{reference image}.
Then, we shift the depth map upwards by 10 pixels to simulate a moving human in a video.
We replace self-attention modules with cross-frame attention as in Text2Video-Zero to enforce the T2I model to generate frames with the same style as the reference frame.
\Cref{fig:intro} shows that cross-frame attention successfully preserves the overall style of the frames.
However, the fine details of the robot (shown in the insets) tend to change between frames.
We find that this is caused by the distributional shift in the latent codes that are responsible for generating the character in the scene as shown on the right of \Cref{fig:intro}.

To this end, we propose a zero-shot approach for generating consistent videos of animated characters based on T2I diffusion models.
To produce continuous motion dynamics, we employ text-based human motion diffusion models \cite{mdm} to generate a sequence of SMPL models given a text prompt.
We render these SMPL models to generate a sequence of depth maps that can be used as conditional inputs for ControlNet.
This allows generating videos with realistic and continuous animations, unlike Text2Video-Zero.
To boost temporal consistency, we compute cross-frame dense correspondence based on DensePose \cite{densepose}, and we use it to align the latent codes between video frames through our \emph{Spatial Latent Alignment} module.
We also propose an additional \emph{Pixel-Wise Guidance} strategy that steers the diffusion process in a direction that minimizes the visual discrepancies between frames.

To evaluate the temporal consistency of the generated videos, we introduce the \emph{Human Mean Squared Error} metric that measures the pixel-wise difference of the animated character between consecutive frames.
Our proposed approach outperforms Text2Video-Zero on this metric by $\sim 10\%$ and was preferred by {76\%} of the users in a user study that we conducted.

\noindent Our contributions can be summarized as follows:
\begin{itemize}
    \item We introduce a zero-shot approach for generating videos of animated characters.
    \item We employ Motion Diffusion Models to generate continuous motion guidance based solely on text.
    \item  We propose the \emph{Spatial Latent Alignment} and \emph{Pixel-Wise Guidance} modules that boost temporal consistency.
    \item Our approach outperforms existing zero-shot approaches in terms of the \emph{Human Mean Squared Error} metric that we introduce and in terms of user preference.
\end{itemize}

%% file: sec/2_related.tex
\section{Related Work}
\label{sec:related}
We give a brief overview of existing approaches for human video synthesis, Text-to-Video (T2V) diffusion models, and human motion synthesis.

\noindent \textbf{Human Video Synthesis}
Existing approaches for human video generation are generally limited to specific domains and datasets.
For instance, several T2V approaches \cite{luo2023videofusion,yu2023magvit,makeavideo} train on the UCF-101 dataset \cite{soomro2012ucf101} that includes videos of humans performing 101 diverse actions.
However, the generated videos based on this dataset are low resolution and lack visual diversity.
Another category of approaches focused on generating videos of fashion performers and is trained on fashion datasets \cite{jiang2023text2performer,deepfashion}.
For instance, Text2Performer \cite{jiang2023text2performer} proposed a decomposed human representation into pose and appearance in the latent space of a variational autoencoder.
This representation is used alongside a diffusion-based motion sampler to generate consistent high-resolution videos of fashion performers.
Nevertheless, their approach can only generate videos of performers with standardized motions on a simple background.

Recently, several approaches \cite{ma2023follow,xu2023magicanimate,hu2023animateanyone} proposed diffusion models for Image-to-Video (I2V) to animate a human character given a subject image and a sequence of poses that are provided by the user.
Contrarily, we address the Text-to-Video (T2V) problem that aims to produce diverse videos of animated characters based solely on a textual prompt.
It is worth mentioning that the concurrent work \cite{cai2023generative} shares similarities with our work as it attempts to generate consistent videos given a sequence of UV maps by aligning the latent codes.
But we differ from them in that we only require textual prompts as input and that we follow a different strategy for aligning the latents.

\noindent \textbf{Text-to-Video Diffusion Models}
Text-to-Image (T2I) diffusion models \cite{imagen,dalle2,sd} excelled in generating highly realistic and diverse images based on textual prompts by harnessing large-scale image datasets \cite{laion5b}.
With the lack of similarly large video datasets to train T2V counterparts, a growing direction of research attempts to exploit existing T2I models to generate videos.
VideoLDM \cite{videoldm} proposed to transform a pre-trained Stable Diffusion model \cite{sd} into a T2V model by introducing a temporal module and a video upsampler that are trained on video data.
Similarly, Make-a-video \cite{makeavideo} extended DALLE-2 \cite{dalle2} to a T2V model by temporally aligning the decoder and the upsampler on video data.
However, these two approaches require excessive GPU resources and large-scale datasets to train.

Tune-a-Video \cite{tuneavideo} adopts a one-shot paradigm and finetunes a pre-trained T2I model to generate a video given a single video/text pair. 
Nevertheless, this approach \emph{requires a video as an input} in addition to the text prompt, making it more suitable for video editing or Video-to-Video tasks.
Text2Video-Zero \cite{text2video-zero} introduced a purely text-based zero-shot T2V approach that injects motion dynamics into the latents of a T2I diffusion model. 
Their approach exploited the fact that the output of diffusion models varies under any changes to the latent codes to generate variations of the first frame.
However, the generated frames from their approach lack any motion continuity or temporal consistency.
In contrast, our approach employs Motion Diffusion Models \cite{mdm,mld} to generate continuous motion guidance and introduces two strategies for boosting temporal consistency, especially at fine details.

\noindent \textbf{Human Motion Synthesis}
This task aims to produce animated skeletons (standardized poses) of humans conditioned on textual prompts.
Several approaches for human motion synthesis were proposed that benefited from the large datasets for human motions, such as the HumanML3D dataset \cite{humanml} with approximately 15k diverse motions.
T2M \cite{humanml} proposed a two-stage approach that learns a mapping function between text prompt and motion length. 
Afterward, a temporal variational autoencoder generates the motion given the predicted length.
MDM \cite{mdm} employed a diffusion model to learn a conditional mapping between text and motion sequences.
MLD \cite{mld} learns a compact latent representation to train diffusion models in a more efficient manner.
GMD \cite{gmd} incorporated spatial constraints into the motion diffusion process to add more control over the generated motions.
We employ any of these approaches to generate diverse and continuous motion signals to guide a pre-trained T2I diffusion model.

%% file: sec/3_method.tex
\begin{figure*}[!ht]
  \centering
  \includegraphics[width=\textwidth]{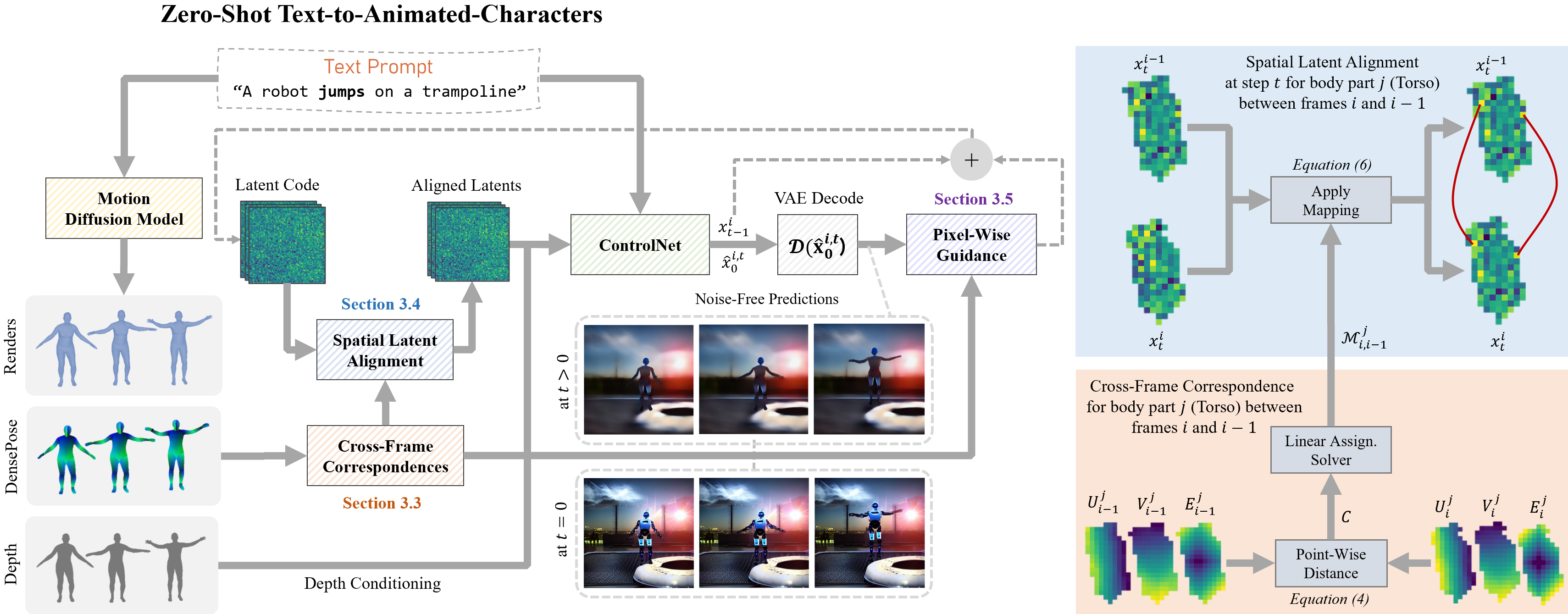}
  \caption{An overview of our proposed approach. Given a text prompt $\mathcal{T}$, a motion diffusion model \cite{mdm} produces a sequence of human skeletons that we use to obtain frame-wise depth maps and DensePose \cite{densepose}. The former is used as guidance for ControlNet \cite{controlnet}, while the latter is used to compute cross-frame correspondences. These correspondences are employed by the Spatial Latent Alignment and the Pixel-Wise Guidance modules to boost temporal consistency. The orange block shows an illustration of how we compute cross-frame correspondences between two frames for the ``torso" body part based on DensePose. The blue block shows how we employ these correspondences to spatially align the latents to promote consistent synthesis.}
  \label{fig:method}
\end{figure*}

\section{Method}
\label{sec:method}

In this section, we first describe the existing pipeline for zero-shot Text-to-Video (T2V) diffusion models that is adopted by Text2Video-Zero \cite{text2video-zero} and MasaCtrl \cite{cao2023masactrl}.
Then, we explain our proposed approach for generating temporally consistent videos of animated characters.

\subsection{Zero-Shot T2V Diffusion Models}
The objective of the T2V task is to generate a sequence of $N$ video frames $\mathcal{I}:= \{ I_1, I_2, \dots, I_N\}$, given a text prompt $\mathcal{T}$.
In the zero-shot setting, a pre-trained Text-to-Image (T2I) diffusion model such as Stable Diffusion (SD) \cite{sd} is used to generate each frame individually.
For better control over the contents of the generated frames, additional conditioning signals $\mathcal{G}:= \{G_1, G_2, \dots, G_N\}$, such as human poses, canny edges, and depth maps are incorporated through ControlNet \cite{controlnet} or T2I-Adapters \cite{t2iadapters}.

During inference, the diffusion process is carried out using a denoising model such as DDIM \cite{ddim}, where for each frame $i$ and denoising step $t$, we compute the previous latent code $x_{t-1}^i$ as well as, a noise-free sample prediction $\xzh$:
\begin{equation}
    \xzh = \dfrac{x_t^i-\sqrt{1-\alpha_t} \ \epsilon_\theta^t(x_t^i, \mathcal{T}, G_i)}{\sqrt{\alpha_t}} \enspace ,
\end{equation}
\begin{equation}
    x_{t-1}^i = \sqrt{\alphatm} \ \xzh + \sqrt{1 - \alphatm - \sigma^2_t} \ \epsilon_\theta^t(x_t^i, \mathcal{T}, G_i) + \sigma_t \epsilon_t \enspace ,
\end{equation}
\noindent where $\alpha_t, \sigma_t$ are pre-defined scheduling parameters, $\epsilon_\theta^t$ is a noise prediction from a trained UNet, and $\epsilon_t$ is random Gaussian noise.
This process is computed for $ T \geq t \geq 0$, and the final image is reconstructed at $t=0$ using the decoder of a variational autoencoder as $I_i = \mathcal{D}(\hat{x}_{0}^{i,0})$.
To promote visual consistency, the initial latent code $x_T$ is shared among all frames, and the self-attention modules are replaced with \emph{cross-frame attention}.
We refer the reader to \cite{text2video-zero, cao2023masactrl} for details on cross-frame attention.
We use this aforementioned pipeline as a baseline for our approach.


\subsection{Zero-Shot Text-to-Animated-Characters}
To generate videos of animated characters using T2I models, we need conditioning signals $\mathcal{G}$ to control the generated content.
Existing methods \cite{tuneavideo,text2video-zero} extract these signals from a user-provided video.
For example, a depth or human pose detector is applied to a video to extract depth maps or human poses.
However, this approach has limited control over the generated content and adds the burden of finding a suitable video.

Instead, we propose to employ text-based motion diffusion models \cite{mdm,mld} to produce a sequence of length $N$ of human skeletons, given the text prompt $\mathcal{T}$.
Afterward, we fit a customizable human body model such as SMPL \cite{smpl} to each of these skeletons, and we render $N$ depth maps from these models to produce conditioning signals $\mathcal{G}^{depth}$.
We also compute DensePose \cite{densepose} for each frame to obtain ${\mathcal{P}:=\{P_1, P_2, \dots, P_N\}}$.
This approach eliminates the need for providing a reference video as in \cite{tuneavideo,text2video-zero,cao2023masactrl} and makes the process purely text-driven.
An overview of our proposed approach is illustrated in Figure \ref{fig:method}.


\subsection{Cross-Frame Dense Correspondences} \label{sec:denspose}
Since we obtained per-frame conditioning signals $\mathcal{G}^{depth}$, we can directly generate the video frames.
However, as demonstrated in \Cref{fig:intro}, the output of SD varies under any changes to the conditioning signal, causing the frames to be temporally inconsistent.
To alleviate this problem, the latent codes during inference must be spatially aligned, \ie, each body part of the generated character needs to have the same latent code in all frames.
To achieve this, we need to compute pixel-wise dense correspondences between frames and use them to propagate the latent codes across frames.

Ideally, the UV maps for the SMPL models or DensePose can be used for this purpose.
However, since they need to be downsampled to the resolution of the latent code, the correspondences are lost, and they need to be re-computed.
To tackle this issue, we set up a dense correspondence problem between each two consecutive frames based on the DensePose embeddings.
We opt for DensePose rather than the UV maps as the former divides the human body into parts, making the correspondence problem cheaper to solve.
We denote the DensePose embedding for frame $i$ as ${P_i=[L_i, U_i, V_i]}$, where $P_i \in \mathcal{P}$, $L_i$ has pixel-wise labels for body parts in the range of $[0,24]$, and $U_i,V_i$ are UV-coordinates in the range of $[0,255]$.
For each body part $j$, we define a set of pixels that belong to that part as ${Q_{i}^{j}:=\{q \ | \ L_i(q)=j\}}$.
We form a feature vector for each $q \in Q_{i}^{j}$ and we arrange them in the rows of matrix $\hat{P}_i^j$:
\begin{equation}
    \hat{P}_i^j[q] = \begin{bmatrix} U_i^j(q) & V_i^j(q) & E_i^j(q) \end{bmatrix} \enspace ,
\end{equation}
where $E_i^j[q]$ is the euclidean distance between pixel $q$ and the centroid of body part $j$.
This term encourages the matching of pixels that are spatially close.

\noindent For each two consecutive frames $i$ and $i-1$, we compute a cost matrix $C$ between $\hat{P}_{i}^j$ and $\hat{P}_{i-1}^j$ as:
\begin{equation}
    C[q,s] = \parallel \hat{P}_{i}^j[q] - \hat{P}_{i-1}^j[s] \parallel_2 \enspace ,
\end{equation}
where $q \in Q_i^j, s \in S_{i-1}^{j}$, and ${S_{i-1}^{j}:=\{s \ | \ L_{i-1}(s)=j\}}$.
Then, we find the correspondences by solving a linear assignment problem over $C$ using 
the Hungarian algorithm \cite{kuhn1955hungarian}, which assigns each pixel to the closest match based on the UV coordinates and the spatial location.
This produces an \emph{injective} mapping for each body part $j$ between frames $i$ and $i-1$ as:
\begin{equation}
    \mathcal{M}_{i,i-1}^j:=\{(q,s) \ \forall \ q\in Q_i^j, s \in S_{i-1}^j\} \enspace ,
\end{equation}
An illustration for this procedure is shown in \Cref{fig:method}.
Finally, All body parts are then combined into a total body mapping as ${\mathcal{M}_{i,i-1} = \cup_j \ \mathcal{M}_{i,i-1}^j}$.


\begin{algorithm}[t]
\caption{Zero-Shot Animated Characters Synthesis}\label{alg:method}
\begin{algorithmic}
    \Require $N,T \in \mathbb{N}$, $\delta \in \mathbb{R}$, text prompt $\mathcal{T}$, $A:=[a_1,a_2]$, $B:=[b_1,b_2]$ ControlNet ($\mathtt{CN}$), DDIM ($\mathtt{DDIM}$), Motion Diffusion Model ($\texttt{MDM}$), Spatial Latent Alignment ($\texttt{ALIGN}$), Pixel-Wise Refinement  ($\texttt{REFINE}$)
    \Ensure   $\mathcal{I}:= \{ I_1, I_2, \dots, I_N\}$
    \State $\mathcal{G}^{depth}, \mathcal{P} \gets \mathtt{MDM}(\mathcal{T})$ \Comment{\small Depth maps, DensePose}
    
    \State $x_T \sim \mathcal{N}(0, I)$
    \For{$i = 1, \dots, N$}:
        \For{$t = T, T-1, \dots, 0$}:
        \If{$i>1$ and $t \in A$} \Comment{\small {Spatial Latent Alignment}}
            \State{$x_t^i \gets \texttt{ALIGN}(x_t^i, \mathcal{P}[i], \mathcal{P}[i-1])$}
        \EndIf
        \State{$\epsilon_\theta^t \gets \texttt{CN}(x_t^i, \mathcal{T}, \mathcal{G}^{depth}[i], t)$}
        \State{$x_{t-1}^i, \xzh \gets \texttt{DDIM}(x_t^i, \epsilon_\theta^t)$}
            \If{$i>1$ and $t \in B$} \Comment{\small {Pixel-Wise Guidance}}
            \State{$\omega_i \gets \texttt{REFINE}(\xzh, \mathcal{P}[i], \mathcal{P}[i-1])$}
            \State{$x_{t-1}^i \gets x_{t-1}^i - \delta \ \nabla_{x_t^i} \ \omega_i$}
            \EndIf
        \EndFor
    \EndFor
\end{algorithmic}
\end{algorithm}


\subsection{Spatial Latent Alignment}
\label{sec:lsa}
To achieve temporal consistency, we aim to align the latents between the video frames.
We compute correspondence mappings from \Cref{sec:denspose} between each two consecutive frames based on DensePose embeddings $\mathcal{P}$ that are downsampled to $64 \times 64$ to match the resolution of the latent codes.
For frames $i$ and $i-1$, the latent code $x_t^i$ is updated with values from $x_t^{i-1}$ based on the computed mapping as:
\begin{equation}
    x_{t}^{i}[q] = x_{t}^{i-1}[s] \quad \forall \ (q,s) \in \mathcal{M}_{i,i-1} \enspace ,
\end{equation}
This operation will copy some parts of the latent code from frame $i-1$ to the correct spatial location in frame $i$, promoting temporal consistency.
Note that we only apply this operation at the first $40 \%$ of the diffusion steps that encompass the generation of the main structures of the scene.


\setlength{\tabcolsep}{3pt}
\begin{figure*}[!ht]    
    \includegraphics[width=\textwidth]{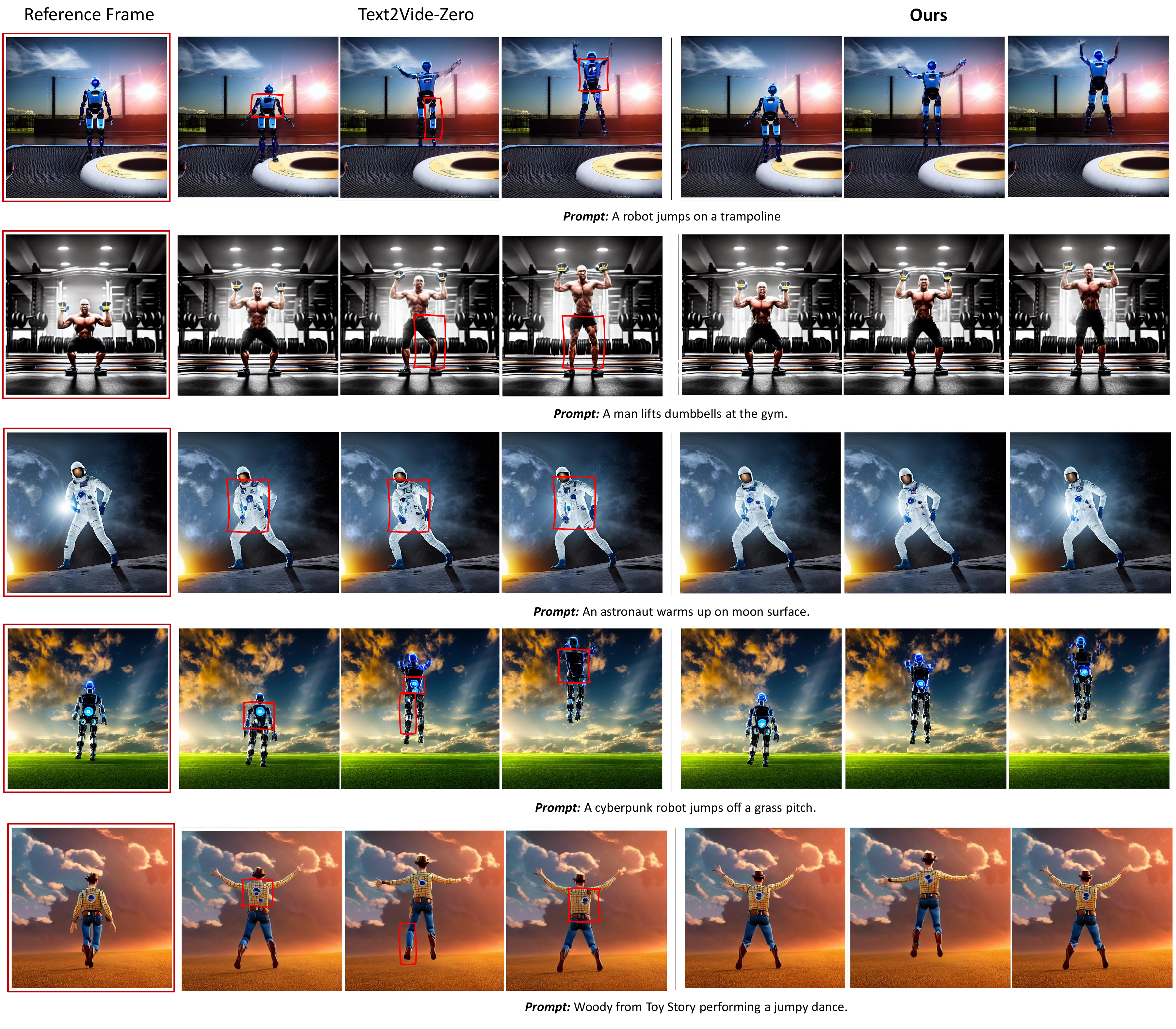}
    \caption{A qualitative comparison between our proposed approach and the baseline Text2Video-Zero \cite{text2video-zero}. Our approach is able to generate consistent shapes and textures compared to the baseline. The reference frame is the first frame of the video that defines the appearance of the character.}
    \label{fig:qual}
\end{figure*}


\subsection{Pixel-Wise Guidance}\label{sec:guidance}
The resolution of the latents in SD is ${1}/{8}$ of that of the generated images.
Consequently, even after spatially aligning the latents in the \Cref{sec:lsa}, some high-resolution details will vary between the video frames.
To alleviate this problem, we propose a Pixel-Wise Guidance strategy inspired by classifier guidance in diffusion models \cite{nichol2021glide}.
First, we compute a mapping $\mathcal{M}_{i,i-1}$ from \Cref{sec:denspose} between each two consecutive frames $i$ and $i-1$.
At a given diffusion step $t$, we reconstruct the RGB predictions using the VAE decoder as $X^i_t = \mathcal{D}(\xzh)$ and we compute the L2 difference between all pixel pairs in $\mathcal{M}^{i,i-1}$:
\begin{equation}
    \omega_i = \sum_{q,s} ( X^i_t[q] - X^{i-1}_t[s] )^2 \quad \forall \ (q,s) \in \mathcal{M}_{i,i-1} \enspace ,
\end{equation}

\noindent Finally, we compute the gradient of $\omega_i$ with respect to $x_t^i$, and we use it to update $x_{t-1}^i$:
\begin{equation} \label{eq:guidance}
    x_{t-1}^i = x_{t-1}^i - \delta \ \nabla_{x_t^i} \ \omega_i \enspace .
\end{equation}
where $\delta$ is a scaling factor. 
This steers the diffusion process in the direction that minimizes $w_i$. 

\noindent Note that we apply this process on the resolution of $256 \times 256$ rather than the full resolution of $512 \times 512$ as the latter would be computationally expensive using the Hungarian algorithm with cubic complexity.


%% file: sec/4_exp.tex
\vspace{-5pt}
\section{Experiments}
\label{sec:exp}



We evaluate our approach based on two baselines that adopt cross-frame attention.
The first baseline is MasaCtrl \cite{cao2023masactrl}, which is an image editing method that can be used to generate a sequence of consistent images, and the second is Text2Video-Zero \cite{text2video-zero}, a zero-shot approach for video synthesis.
Note that Text2Video-Zero has two variations: a condition-free version and a conditional one based on ControlNet.
We compare mainly against the latter, but we provide some examples for the former as well.

\subsection{Implementation Details}
For both baselines, we use a pre-trained Stable Diffusion \cite{sd} version 1.5 with ControlNet \cite{controlnet} depth control to generate $512 \times 512$ video frames.
For inference, we employ the DDIM sampler \cite{ddim} with a linear schedule.
We use $T=100$ inference steps for Text2Video-Zero and $T=50$ for MasaCtrl.
We empirically choose the guidance factor ${\delta=0.01}$ in \Cref{eq:guidance}, $A=[0,39], B=[20,69]$ for Text2Video-Zero, and $A=[0,19], B=[20,39]$ for MasaCtrl in \Cref{alg:method}.
For motion synthesis, we use the official implementation of \emph{Motion Diffusion Model (MDM)} \cite{mdm} with some modifications for rendering and computing DensePose \cite{densepose}.
We conduct all experiments on a single NVIDIA A100 GPU except for Gen-2 \cite{gen1}, where we use the official demo.
The code is publicly available \footnote{\url{https://github.com/abdo-eldesokey/latentman}}.

\subsection{Qualitative Results}
\setlength{\tabcolsep}{1pt}
\begin{figure}[!t]
     \begin{tabular}{cc | cc}
        \multicolumn{2}{c}{Text2Video-Zero \cite{text2video-zero}} & \multicolumn{2}{c}{\textbf{Ours}} \\
        \includegraphics[width=0.24\columnwidth,trim={0, 0, 0, 0}, clip]{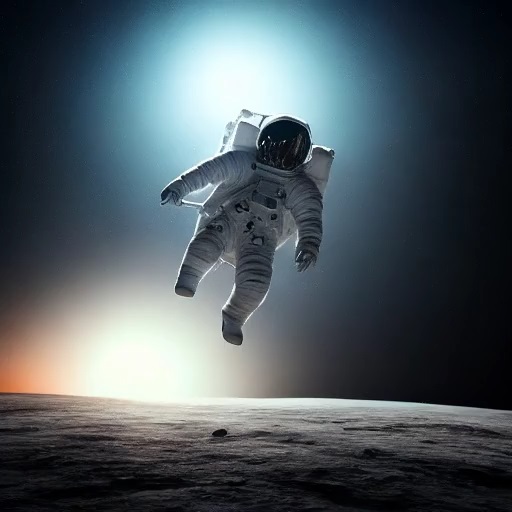} &  
        \includegraphics[width=0.24\columnwidth,trim={0, 0, 0, 0}, clip]{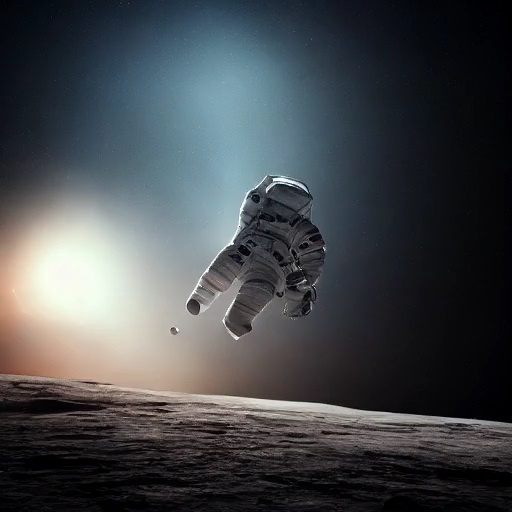} &          
        \includegraphics[width=0.24\columnwidth,trim={0, 0, 0, 0}, clip]{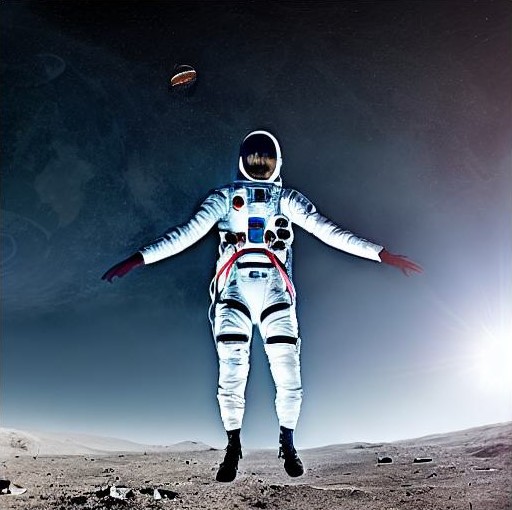} &  
        \includegraphics[width=0.24\columnwidth,trim={0, 0, 0, 0}, clip]{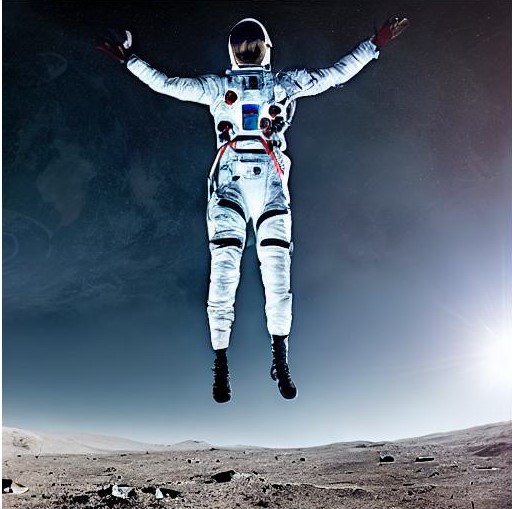} \\
        \multicolumn{4}{c}{\prompt{An astronaut jumps on the moon surface}} \\
        \includegraphics[width=0.24\columnwidth,trim={0, 0, 0, 0}, clip]{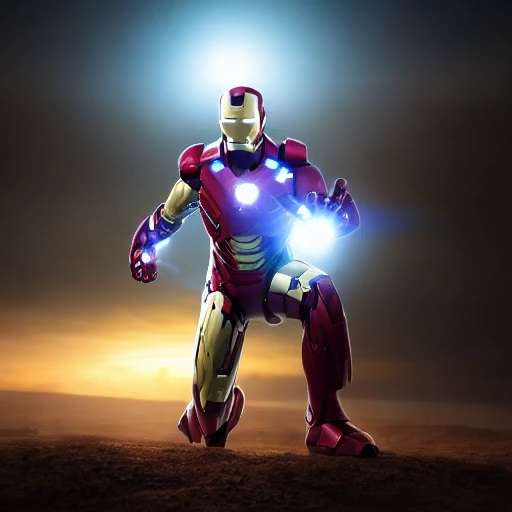} &  
        \includegraphics[width=0.24\columnwidth,trim={0, 0, 0, 0}, clip]{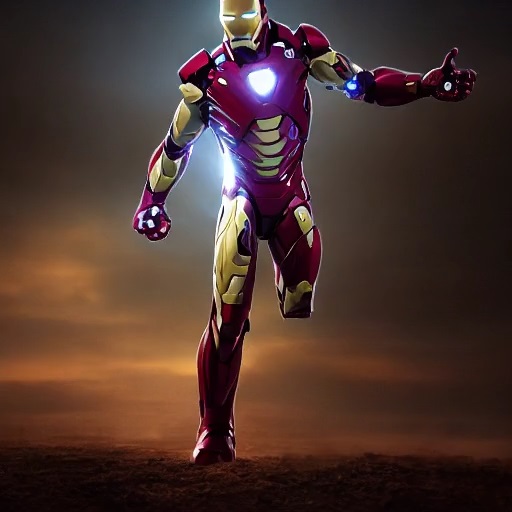} &          
        \includegraphics[width=0.24\columnwidth,trim={30px, 20px, 30px, 40px}, clip]{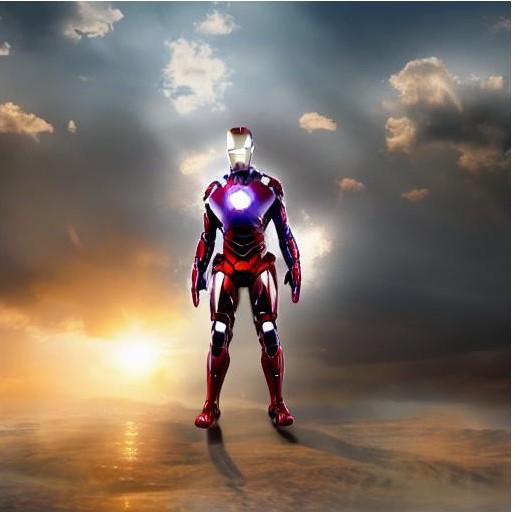} &  
        \includegraphics[width=0.24\columnwidth,trim={30px, 20px, 30px, 40px}, clip]{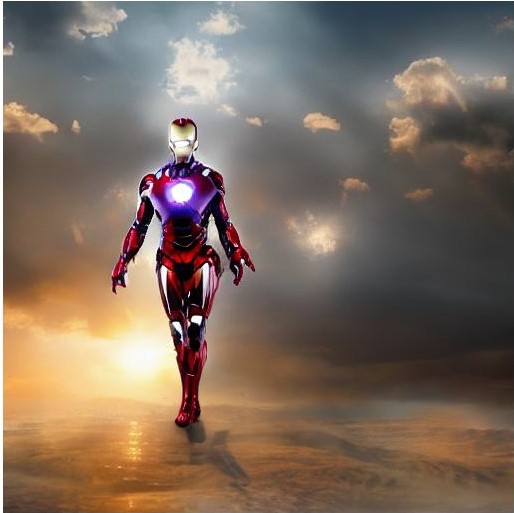} \\
        \multicolumn{4}{c}{\prompt{Ironman moves to the right}}\\
    \end{tabular}
    \caption{Impact of motion guidance in our approach compared to the motion dynamics in Text2Video-Zero \cite{text2video-zero}. Motion guidance produces videos that adhere to the prompt in contrast to Text2Video-Zero, which produces random variations of the scene.}
    \label{fig:no_guidance}
\end{figure}

\textbf{Zero-Shot Comparison} \Cref{fig:qual} shows a qualitative comparison between depth-conditioned Text2Video-Zero and our proposed approach.
The figure shows that cross-frame attention adopted by Text2Video-Zero is not sufficient to preserve the fine details of the generated characters.
As the conditional depth map changes, the object gets distorted (\eg the robot torso and legs in the first row), or the texture changes (\eg the pants in the second row become shorts).
On the other hand, our approach successfully maintains the fine details of the generated characters across all frames.

\noindent \textbf{Motion Guidance Significance} To demonstrate the impact of motion guidance produced by MDM, we compare our approach against Text2Video-Zero with no depth conditioning, which produces video by injecting motion dynamics into the latent codes.
\Cref{fig:no_guidance} shows that Text2Video-Zero produces random variations of the scene that do not adhere to the motion in the prompt.
For example, the astronaut in the top row is just floating and not jumping, and Ironman in the second row does not move but changes pose.
On the other hand, our approach produces consistent videos that adhere to the motion in the prompt.

\begin{figure*}[!t]
     \begin{tabular}{c ccc c | c ccc}
        \rotatebox{90}{ \ \ Gen-2 \cite{gen1}} & \includegraphics[width=0.158\textwidth, trim={100, 0, 100, 0}, clip]{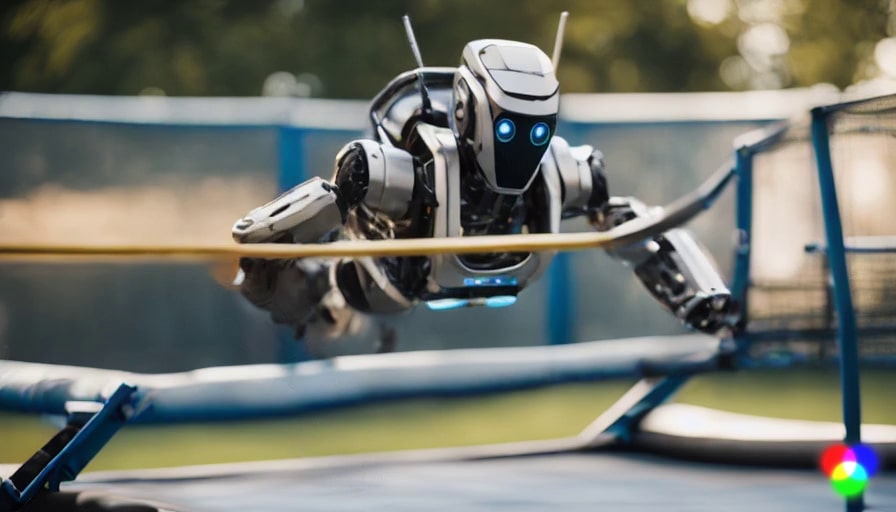} &  
        \includegraphics[width=0.158\textwidth,trim={100, 0, 100, 0}, clip]{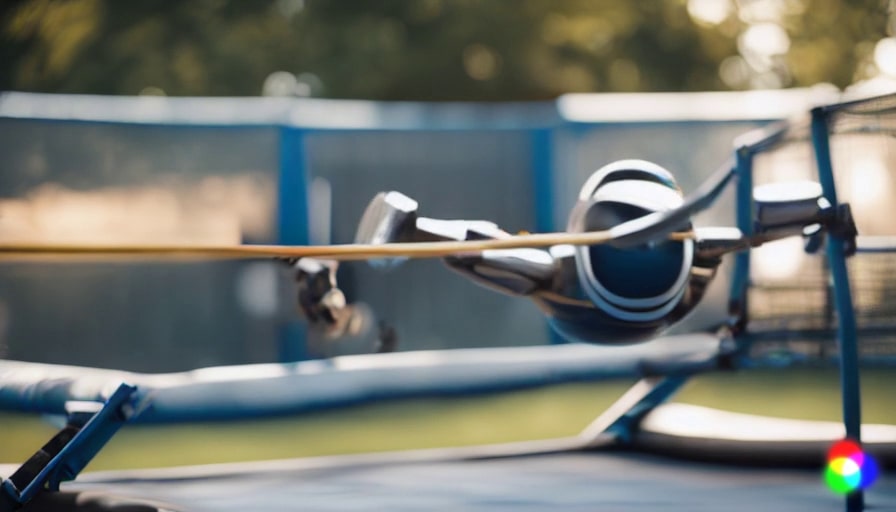} &  
        \includegraphics[width=0.158\textwidth,trim={100, 0, 100, 0}, clip]{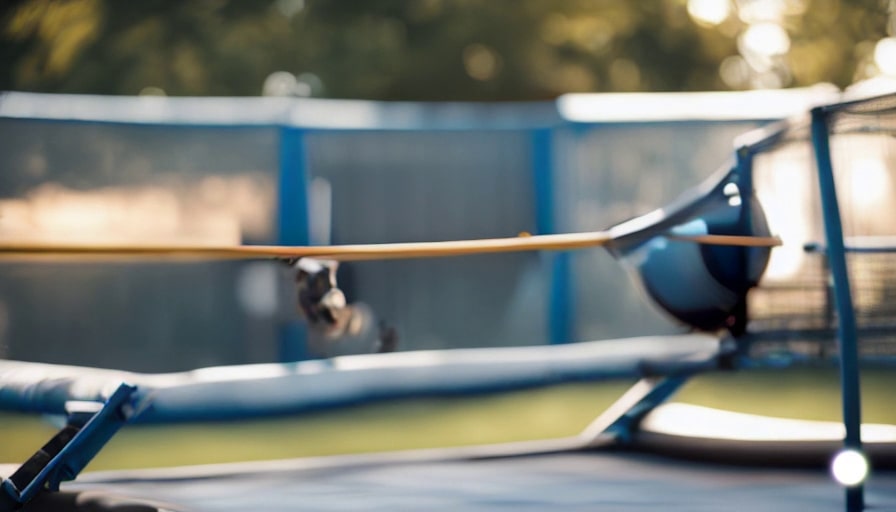} & & &
         \includegraphics[width=0.158\textwidth, trim={200, 0, 200, 0}, clip]{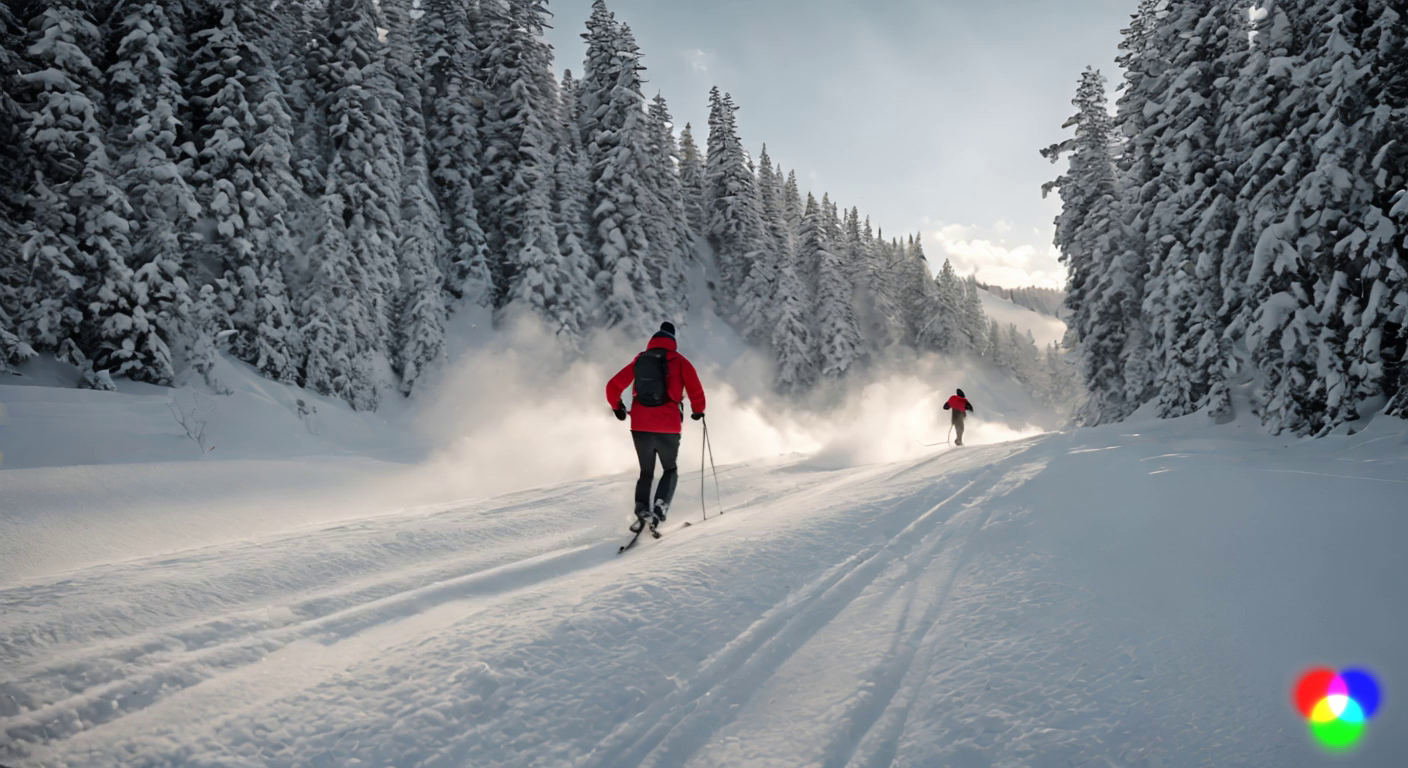} & 
         \includegraphics[width=0.158\textwidth,trim={200, 0, 200, 0}, clip]{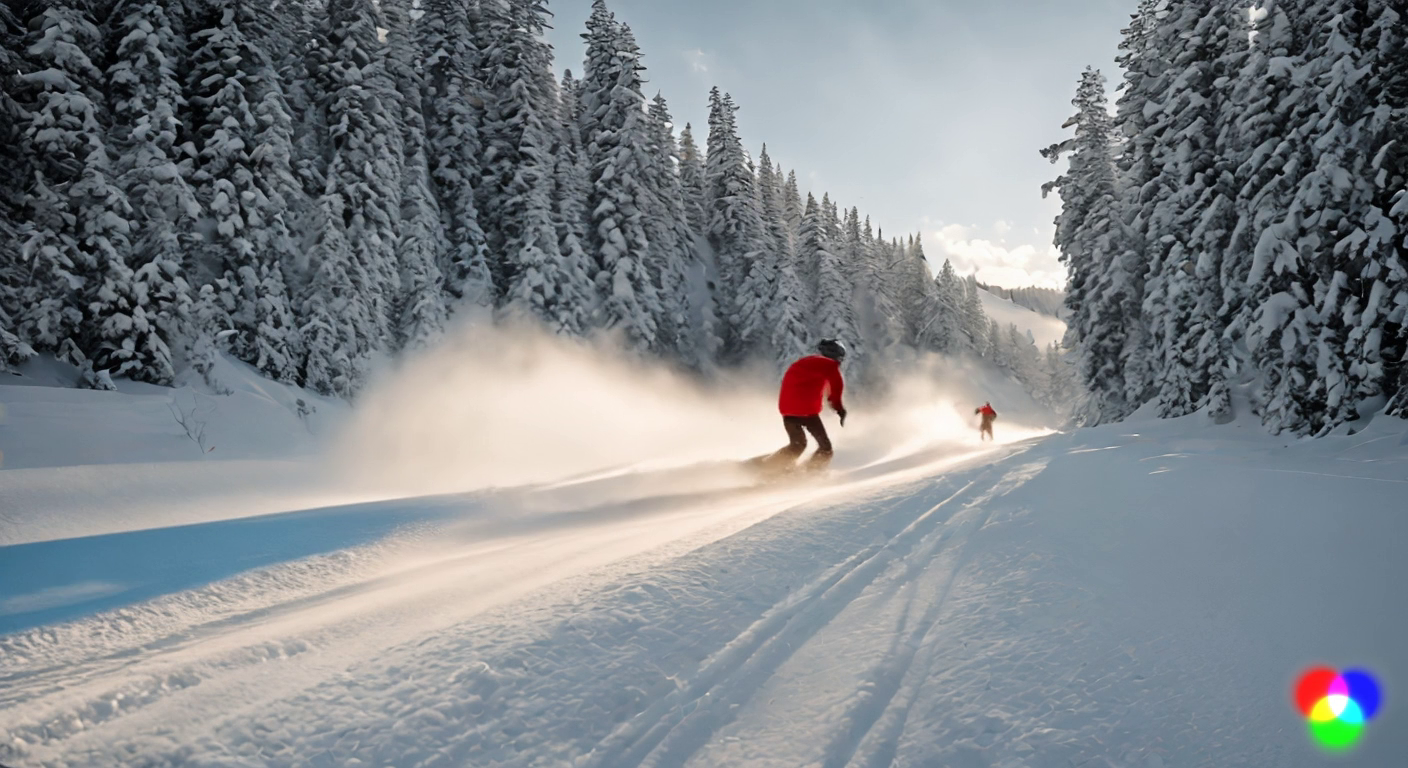} &  
        \includegraphics[width=0.158\textwidth,trim={200, 0, 200, 0}, clip]{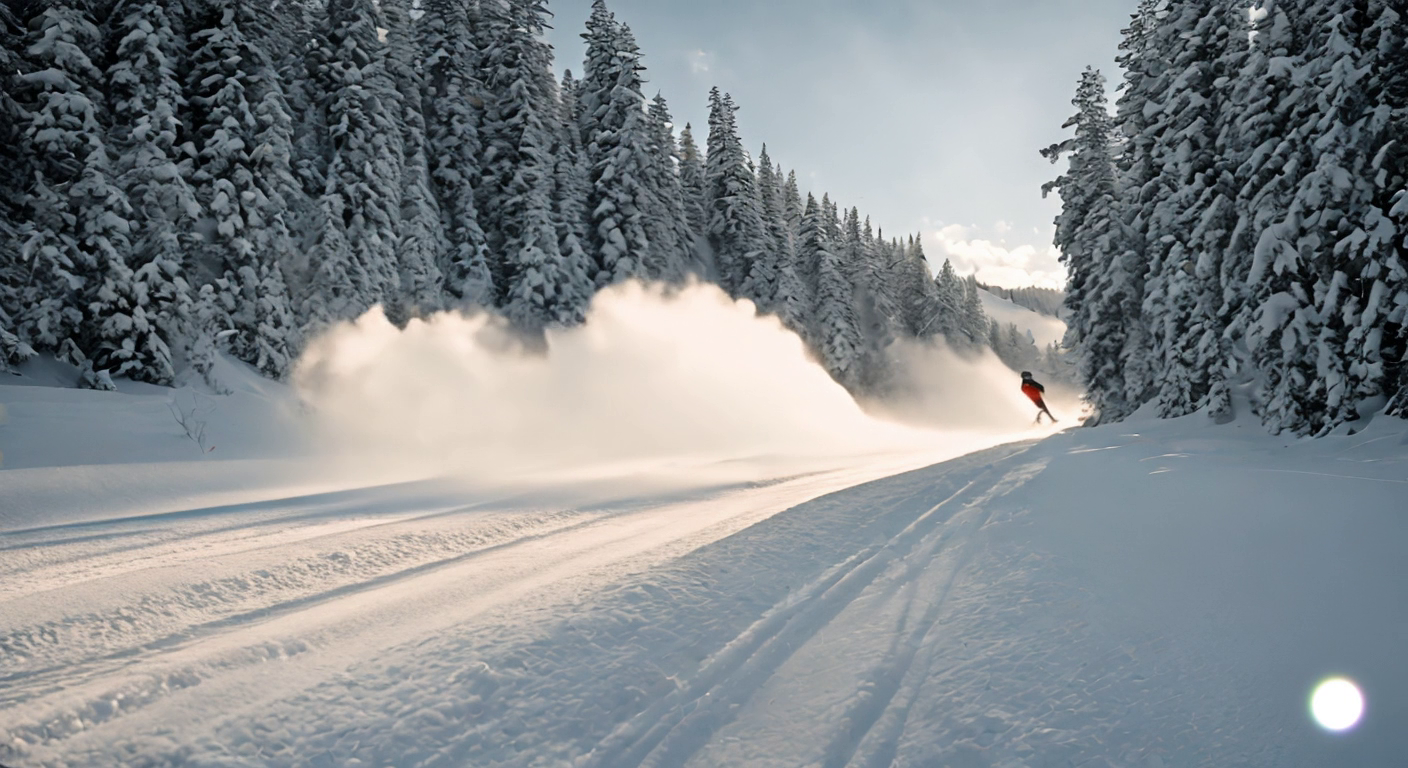}  \\
        
        \rotatebox{90}{\qquad \textbf{Ours}} & 
        \includegraphics[width=0.15\textwidth, trim={0, 90, 0, 0}, clip]{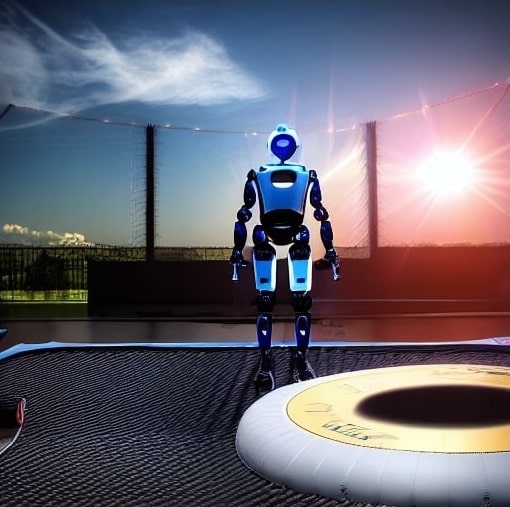} &   
        \includegraphics[width=0.15\textwidth, trim={00, 90, 0, 0}, clip]{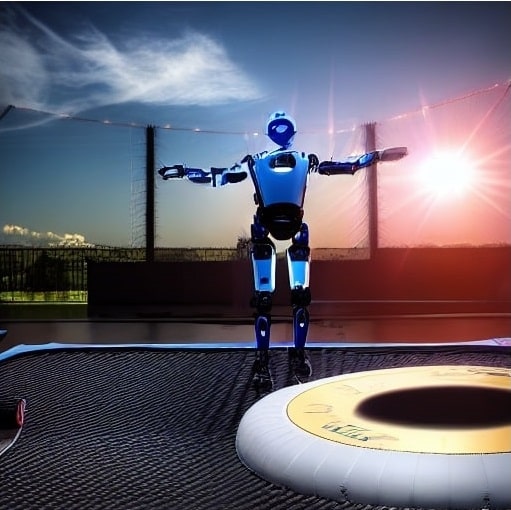} &   
        \includegraphics[width=0.15\textwidth, trim={0, 90, 0, 0}, clip]{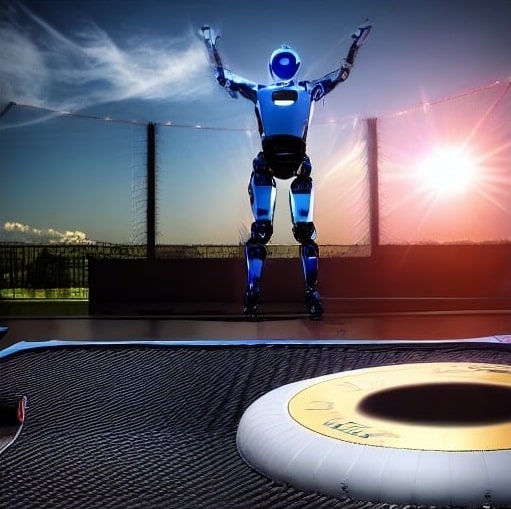} &&&
        \includegraphics[width=0.15\textwidth, trim={0, 0, 0, 90}, clip]{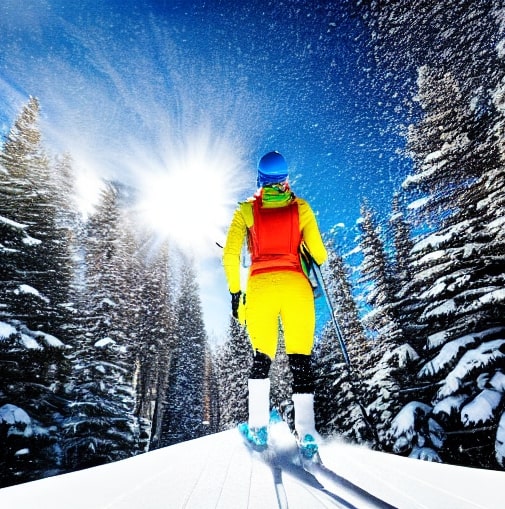} &   
        \includegraphics[width=0.15\textwidth, trim={0, 0, 0, 90}, clip]{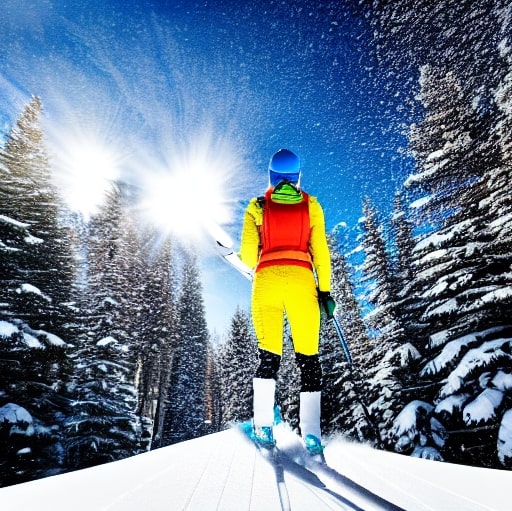} &   
        \includegraphics[width=0.15\textwidth, trim={0, 0, 0, 90}, clip]{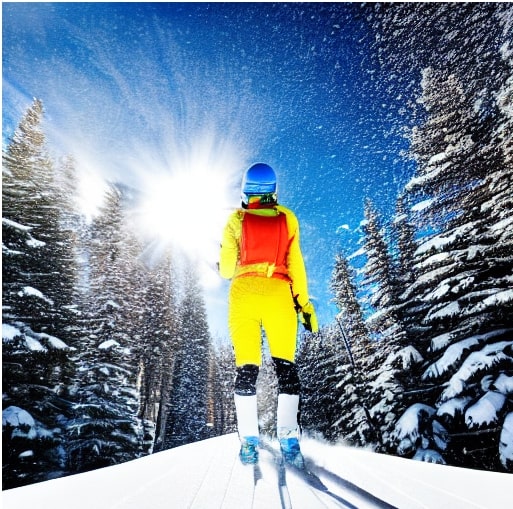} \\
    
        &\multicolumn{3}{c}{\prompt{A robot jumps on a trampoline}} &&& \multicolumn{3}{c}{\prompt{A skier running on a snowy road}} \\
    \end{tabular}
    \caption{A comparison against the trained T2V model Gen-2 \cite{gen1}.}
    \label{fig:qual_t2v}
\end{figure*}

\noindent \textbf{Trained T2V Comparison} We also provide a comparison against the trained T2V model, Gen-2 \cite{gen1}, in \Cref{fig:qual_t2v}.
In the first column, Gen-2 fails to produce a video of a robot jumping on a trampoline, and the robot morphs into a sphere.
Our approach manages to produce a video for this uncommon scenario as the generation of the motion and the style are decoupled.
In the second column, Gen-2 produces a good video of a skier with rich video dynamics.
However, the skier loses his backpack after a few frames and deforms by the end of the video.
Our approach produces a consistent video but with less background dynamics.


\subsection{Quantitative Results}

\begin{table}
    \centering
    \begin{tabular}{L{4.0cm}|C{2.1cm}|C{2.0cm}}         
         \hline
         & $H_{MSE} \downarrow$ &   \small User Preference [\%] \\
         \specialrule{1pt}{1pt}{1pt}
        MasaCtrl \cite{cao2023masactrl} \tiny[ICCV23] & 88.19 & 34 \% \\
        \hline
        \multirow{1}{*}{\textbf{Ours}} & \textbf{79.88} & \multirow{2}{*}{\textbf{66 \%}} \\
         & \small(-9.4 \%) & \\
        \specialrule{1pt}{1pt}{1pt}        
        Text2Video-Zero \cite{text2video-zero} \tiny[ICCV23] & 84.87 & 24 \%  \\
        \hline
        \multirow{1}{*}{\textbf{Ours}} & \textbf{76.41}  & \multirow{2}{*}{\textbf{76 \%}} \\
         & \small{(-10.0 \%)} & \\        
        \hline
        
    \end{tabular}
    \caption{Quantitative comparison between our proposed approach and two baselines. The error reduction percentage is shown between brackets. }
    \label{tab:quant}
\end{table}

\setlength{\tabcolsep}{1pt}

To numerically evaluate the generated videos, we introduce a new metric for temporal consistency and perform a user study.
We denote the new metric as the \emph{Human Mean Squared Error} $H_{MSE}$, and it compares the pixel-wise values of the generated characters in every two consecutive frames.
We employ the computed cross-frame dense correspondences $\mathcal{M}$ from \Cref{sec:denspose}, and we compute the mean squared error (MSE) between corresponding pixels:
\begin{align}
    H_{MSE} = \dfrac{1}{N} \sum_{i=1}^{N} \dfrac{1}{|\mathcal{M}_{i,i-1}|} \sum_{q,s} (I_i[q] - I_{i-1}[s])^2 \nonumber \\
    \forall \ (q,s) \in \mathcal{M}_{i,i-1} 
\end{align}
where $I_i, I_{i-1}$ are the final generated frames in $\mathcal{I}$.

We generated 10 videos of diverse motions and characters and we computed the proposed metric and performed the user study on them. 
\Cref{tab:quant} shows that our approach outperforms both baselines in terms of $H_{MSE}$ by $\sim 9-10\%$, which demonstrates that the generated characters are temporally more consistent. 
For the user study, users are asked to select between two videos; one is produced by the baseline and the other by our approach.
\Cref{tab:quant} shows that $76 \%$ and $66 \%$ of the users (based on 23 users) preferred the videos generated by our approach over Text2Video-Zero and MasaCtrl baselines, respectively.

\vspace{-5pt}
\subsection{Ablation Study}
We provide an ablation study in \Cref{tab:abl} to show the contribution of each component in our proposed pipeline to the overall performance.
The Spatial Latent Alignment module contributes the most to the overall improvement and improves by $7.0\%$ over the baseline.
This indicates that aligning the latents plays a crucial role in achieving temporal consistency.
Pixel-Wise Guidance in \Cref{sec:guidance} improves over the baseline by $2.6 \%$ as it is mainly focused on the fine details.
The two components combined achieve a joint improvement of 10 \% compared to the baseline.

\begin{table}
    \centering
    \begin{tabular}{L{3cm}|C{3cm}|C{2cm}}
         \hline
         & $H_{MSE} \downarrow$ & Runtime (s) \\
         \specialrule{1pt}{1pt}{1pt}         
        Baseline  &  84.87 & 28 \\
        \hline
        with SLA & 78.90 (-7.0 \%) & 30   \\
        \hline
        with PWG & 82.73 (-2.5 \%) & 49\\
        \hline
        with LSA + PWG  & 76.41 (-10.0 \%) & 50  \\        
        \hline
    \end{tabular}
    \caption{An ablation study for different components of our proposed approach. \emph{SLA}: Spatial Latent Alignment in \Cref{sec:lsa}, \emph{PWG}: Pixel-Wise Guidance in \Cref{sec:guidance}. Runtime is reported for generating a 8-frames video.}
    \label{tab:abl}
\end{table}


\subsection{Limitations and Failure Cases}
\begin{figure}[!t]
     \begin{tabular}[t]{cc  cc}
        \includegraphics[width=0.14\columnwidth, trim={0, 0, 0, 0}, clip]{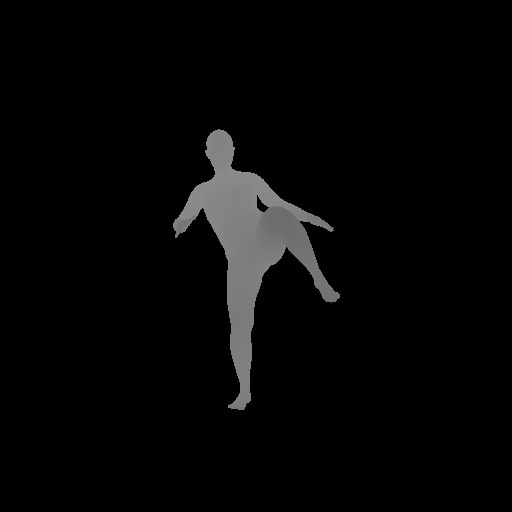} &  \includegraphics[width=0.34\columnwidth, trim={10, 60, 522, 100}, clip]{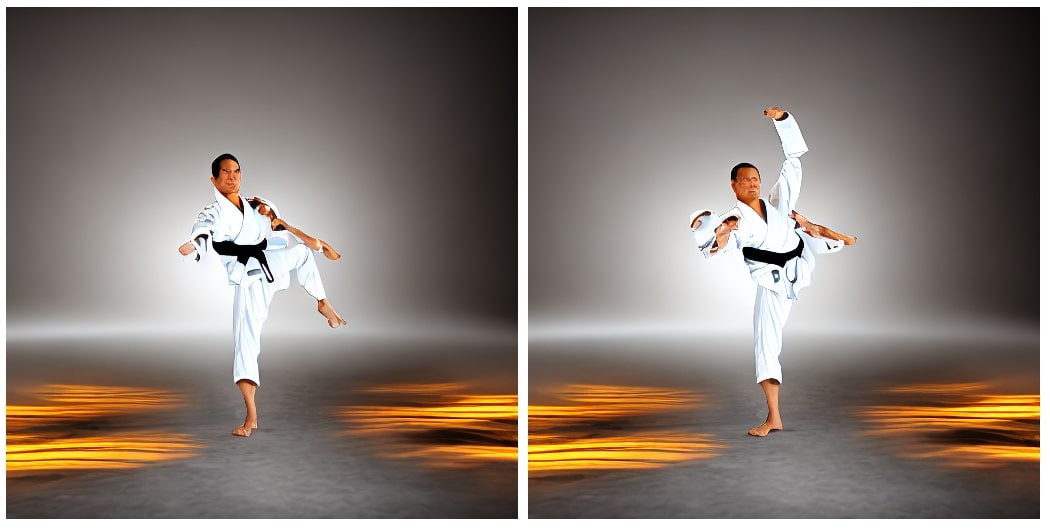} &
        \includegraphics[width=0.14\columnwidth, trim={0, 0, 0, 0}, clip]{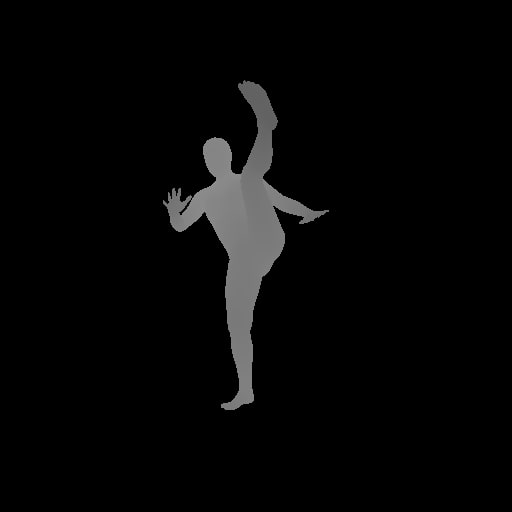} &
        \includegraphics[width=0.34\columnwidth, trim={522, 60, 10, 100}, clip]{fig/fail/karate_fail.jpg} \\
        
    \end{tabular}
    \caption{Examples of a failure case}
    \label{fig:fail}
\end{figure}

Since we employ ControlNet with depth conditioning for generating the video frames, our approach is also bounded by its limitations.
For example, the top row of \Cref{fig:fail} shows an example where ControlNet fails to produce a realistic left arm and leg when they intersect in the depth map.
Another source of failure is mismatches when computing the correspondence mapping in \Cref{sec:denspose}, which can lead to some artifacts.
It is also worth mentioning that Pixel-Wise Guidance imposes high GPU memory usage due to computing gradients with respect to the latent codes.
However, employing the Spatial Latent Alignment solely still achieves remarkable improvement over the baseline as shown in \Cref{tab:abl} with no GPU memory overhead.

\setlength{\tabcolsep}{1pt}


%% file: sec/5_conclusion.tex
\section{Conclusion and Future Work}
We introduced a new paradigm for generating consistent videos of animated characters in a zero-shot manner. 
We employed text-based motion diffusion models to provide continuous motion guidance that we utilized to generate video frames through a pre-trained T2I diffusion model.
This allowed generating videos of diverse characters and motions that existing T2V methods struggled to produce.
We also demonstrated that our approach produces temporally consistent videos achieved through the proposed Spatial Latent Alignment and Pixel-Wide Guidance modules.
These two modules can benefit other approaches that adopt cross-frame attention and latent diffusion models in general.
For future work, the cross-frame dense correspondences can be improved for better latent alignment.
Furthermore, video dynamics can be incorporated into the background for enhanced realism.

%% file: X_suppl.tex
\clearpage
\setcounter{page}{1}
\setcounter{figure}{0}
\setcounter{section}{0}

\maketitlesupplementary
\resetlinenumber

\begin{figure}[!h]
    \centering
    \includegraphics[width=\columnwidth]{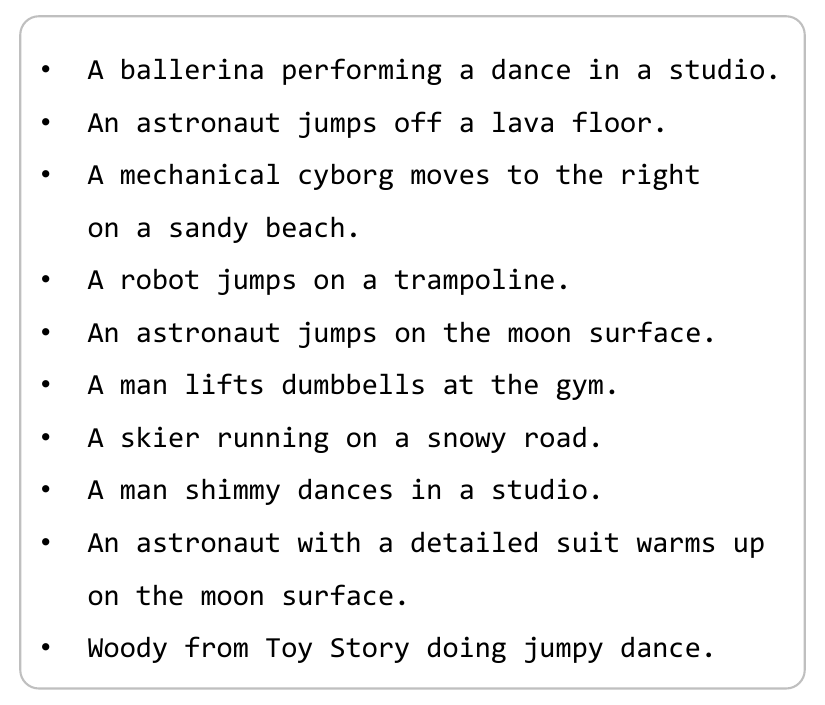}
    \caption{Prompts used to generate videos for the user study.}
    \label{fig:prompts}
\end{figure}
\section{Additional Results}
\begin{figure*}[!ht]    
    \begin{tabular}{c c}                   
        \rotatebox{90}{ \small \qquad MasaCtrl \cite{cao2023masactrl}} & \includegraphics[width=0.96\textwidth, trim={0 0 1576 0},clip]{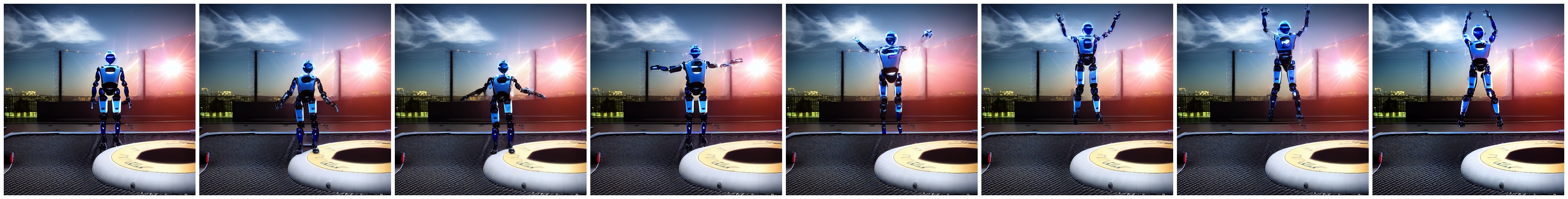} \\
        \rotatebox{90}{ \small  \quad Text2Video-Zero \cite{text2video-zero} } & \includegraphics[width=0.96\textwidth, trim={0 0 1576 0},clip]{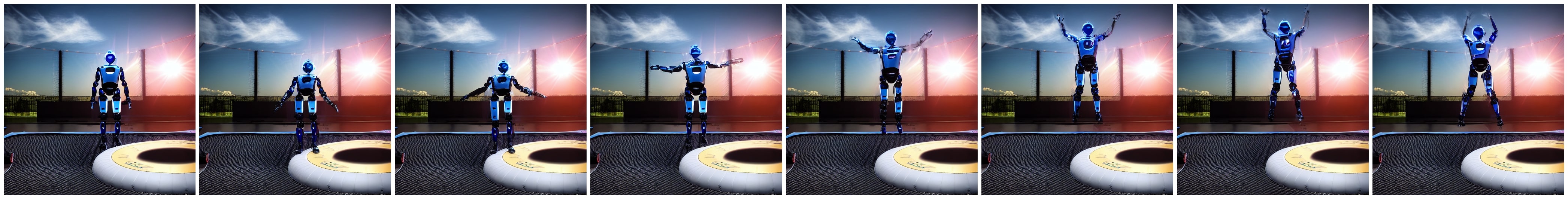} \\
        \rotatebox{90}{ \small  \qquad \qquad \textbf{Ours} } & \includegraphics[width=0.96\textwidth, trim={0 0 1576 0},clip]{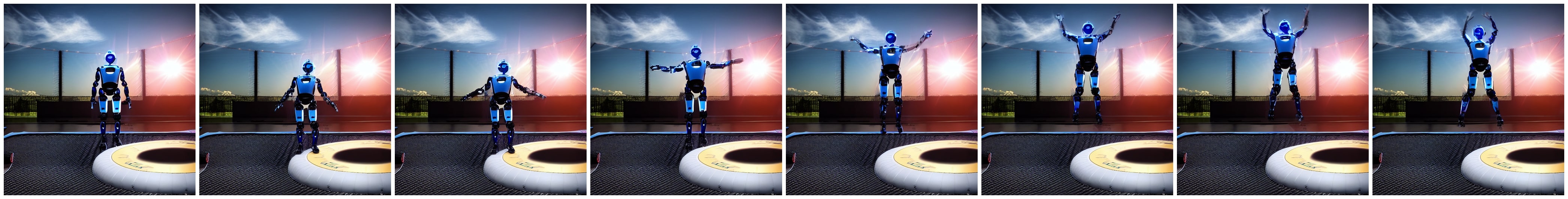} \\        
        \multicolumn{2}{c}{\prompt{A robot jumps on a trampoline}} \\

        \rotatebox{90}{ \small \qquad MasaCtrl \cite{cao2023masactrl}} & \includegraphics[width=0.96\textwidth, trim={1044 0 532 0},clip]{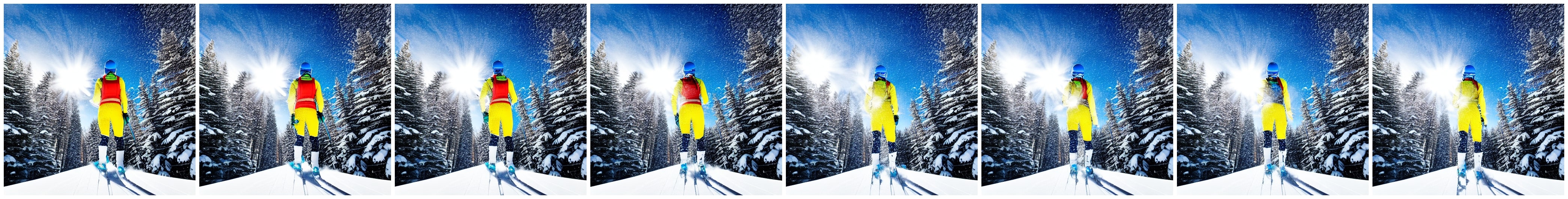} \\
        \rotatebox{90}{ \small  \quad Text2Video-Zero \cite{text2video-zero} } & \includegraphics[width=0.96\textwidth, trim={1044 0 532 0},clip]{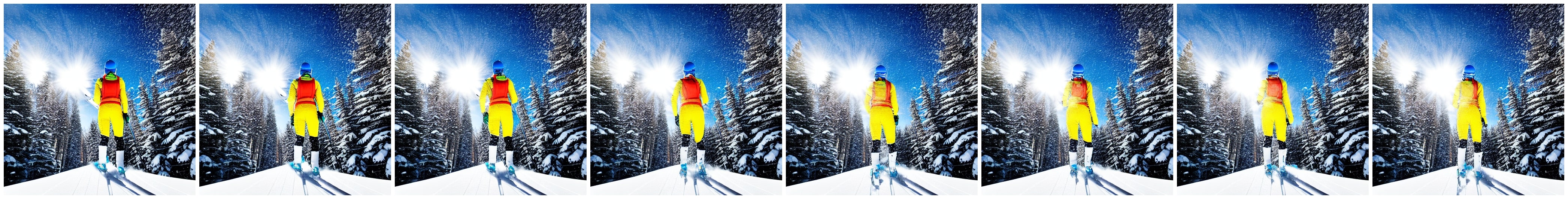} \\
        \rotatebox{90}{ \small  \qquad \qquad \textbf{Ours} } & \includegraphics[width=0.96\textwidth, trim={1044 0 532 0},clip]{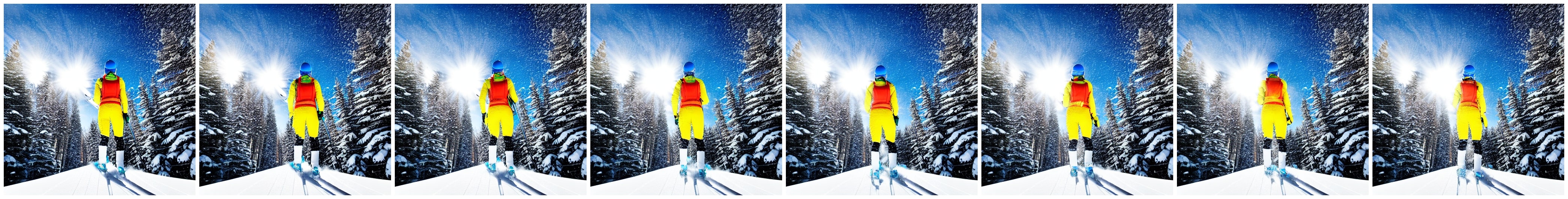} \\        
        \multicolumn{2}{c}{\prompt{A skier running on a snowy road}} \\
        
    \end{tabular}
    \caption{A qualitative comparison between our proposed approach, MasaCtrl \cite{cao2023masactrl}, and Text2Video-Zero \cite{text2video-zero}.}
    \label{fig:qual2}
\end{figure*}

\begin{figure*}[!ht]    
    \begin{tabular}{c c}                   
        \rotatebox{90}{ \small \qquad MasaCtrl \cite{cao2023masactrl}} & \includegraphics[width=0.96\textwidth, trim={1044 0 532 0},clip]{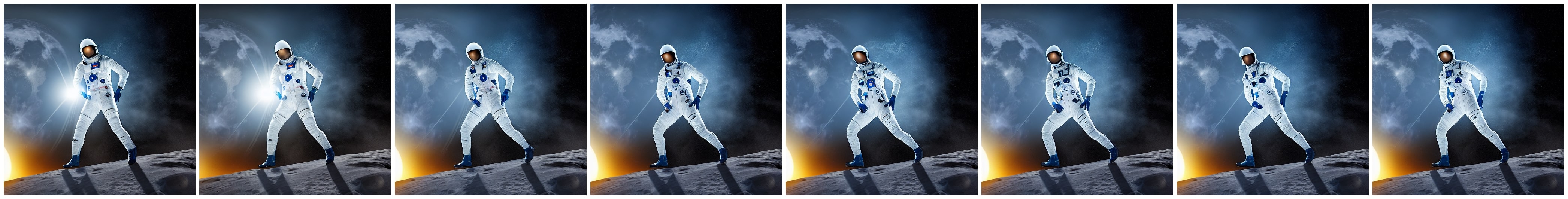} \\
        \rotatebox{90}{ \small  \quad Text2Video-Zero \cite{text2video-zero} } & \includegraphics[width=0.96\textwidth, trim={1044 0 532 0},clip]{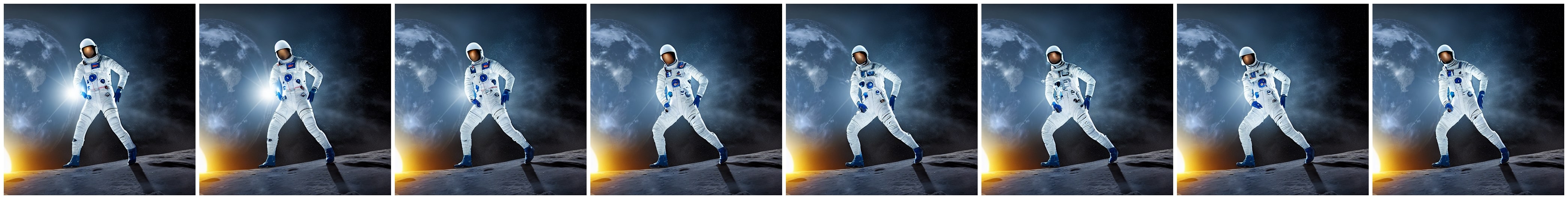} \\
        \rotatebox{90}{ \small  \qquad \qquad \textbf{Ours} } & \includegraphics[width=0.96\textwidth, trim={1044 0 532 0},clip]{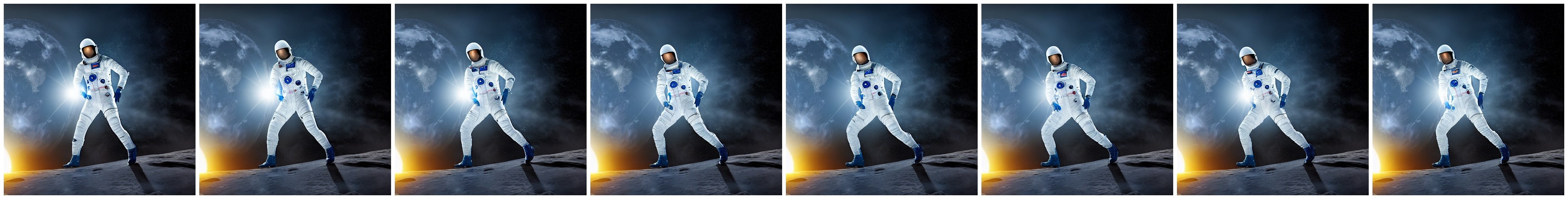} \\      
        \multicolumn{2}{c}{\prompt{An astronaut with detailed suit warms up on the moon surface}} \\

        \rotatebox{90}{ \small \qquad MasaCtrl \cite{cao2023masactrl}} & \includegraphics[width=0.96\textwidth, trim={1044 0 532 0},clip]{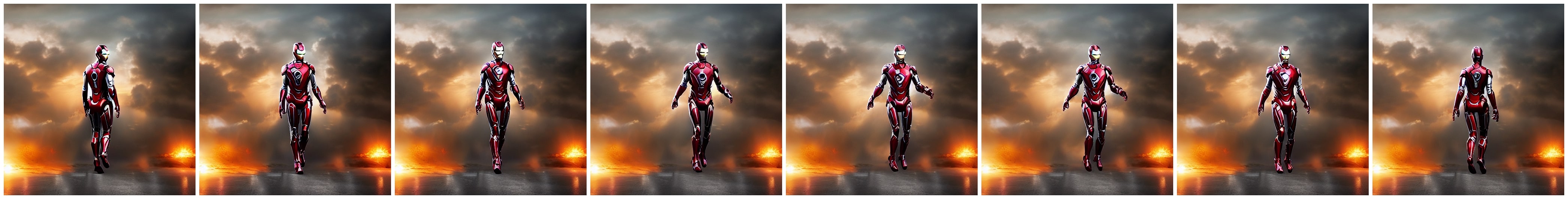} \\
        \rotatebox{90}{ \small  \quad Text2Video-Zero \cite{text2video-zero} } & \includegraphics[width=0.96\textwidth, trim={1044 0 532 0},clip]{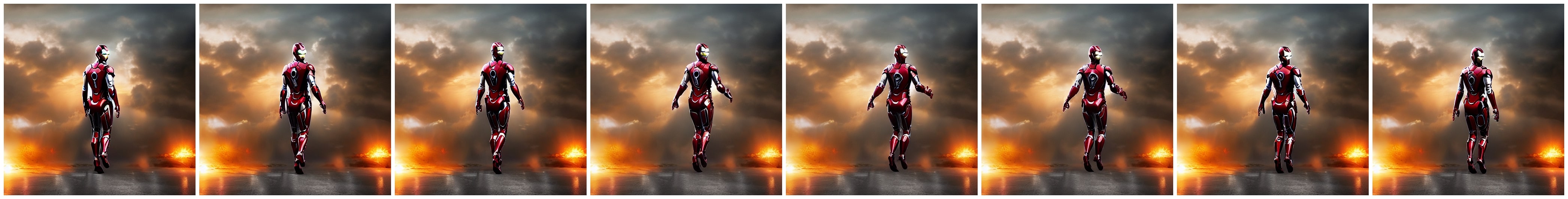} \\
        \rotatebox{90}{ \small  \qquad \qquad \textbf{Ours} } & \includegraphics[width=0.96\textwidth, trim={1044 0 532 0},clip]{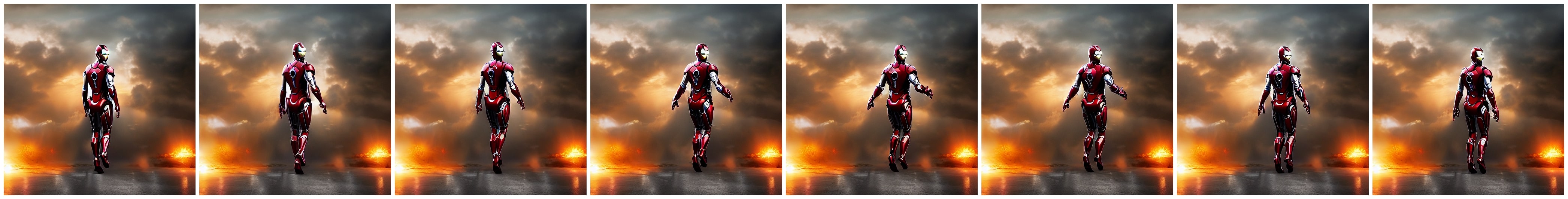} \\      
        \multicolumn{2}{c}{\prompt{Ironman is walking in the street}} \\
        
    \end{tabular}
    \caption{A qualitative comparison between our proposed approach, MasaCtrl \cite{cao2023masactrl},  and Text2Video-Zero \cite{text2video-zero}.}
    \label{fig:qual3}
\end{figure*}

We provide additional \textbf{video} results generated by our method and the two approaches in comparison in the \href{https://abdo-eldesokey.github.io/text2ac-zero/}{project page}.
Additional image results are also shown in \Cref{fig:qual2,fig:qual3}.
Note that we use Stable Diffusion \cite{sd} version 1.5, and the same seed when generating results for different approaches to produce closely comparable videos.


\section{User Study}
For the user study, we generated 10 videos of various motions and objects using Text2Video-Zero \cite{text2video-zero}, MasaCtrl \cite{cao2023masactrl}, and our method.
The prompts used to generate these videos are provided in \Cref{fig:prompts}.
We conducted two separate user studies between our method vs. MasaCtrl and ours vs. Text2Video-Zero. 
Users were asked to select the video where the character has stable/consistent textures throughout the video.


\section{Cross-Frame Correspondences}
\begin{figure*}[!t]
  \centering
  \includegraphics[width=0.8\textwidth]{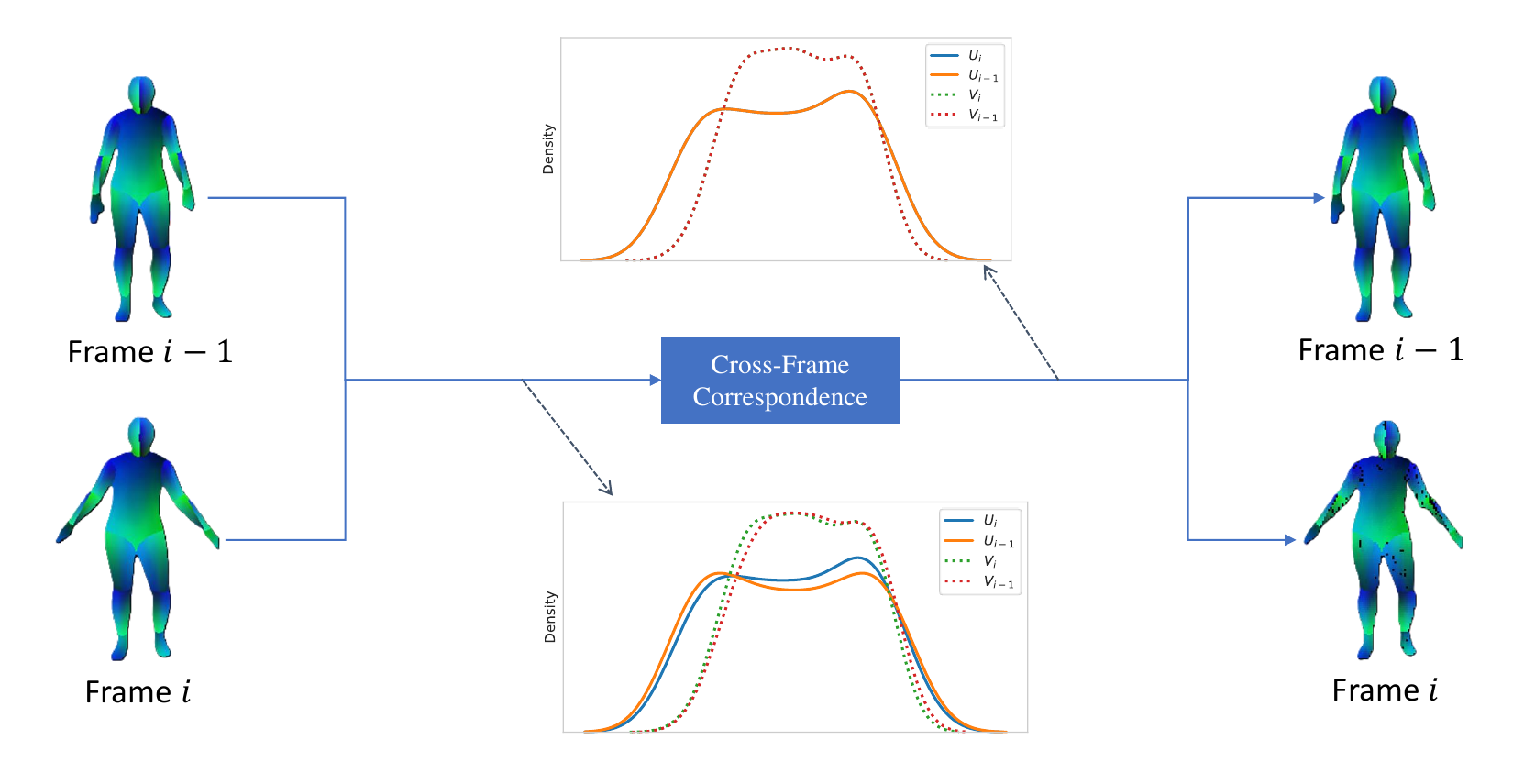}
  \caption{DensePose embeddings for different frames have dissimilar distributions for the UV-coordinates. By computing cross-frame correspondences, we align these coordinates, and we obtain a pixel-wise mapping between the two frames. }
  \label{fig:corr}
\end{figure*}

We explain why there is a need for computing cross-frame correspondences in \Cref{sec:denspose} based on DensePose \cite{densepose}.
\Cref{fig:corr} shows that the DensePose embeddings for two consecutive frames have different distributions for the UV-coordinates.
This makes it difficult to track different body parts across frames.
By computing cross-frame correspondences based on DensePose, the UV-coordinates are matched between frames, and we obtain a pixel-wise mapping.
This allows propagating various signals between frames, such as the latent codes and RGB values.
Note that this mapping is \emph{injective}, meaning that each pixel in frame $i$ is mapped to a \emph{single} or \emph{no} pixels in frame $i-1$.
The unmatched pixels are shown in black in \Cref{fig:corr} for illustration, but in practice, they are replaced with the original values from the DensePose embedding for frame $i$.

\section{Qualitative Ablation Study}
\begin{figure*}[!ht]    
    \begin{tabular}{c|c|c}                   
        Only SLA & Only PWG  & SLA + PWG\\
        \includegraphics[width=0.32\textwidth, trim={2098 0 1054 0},clip]{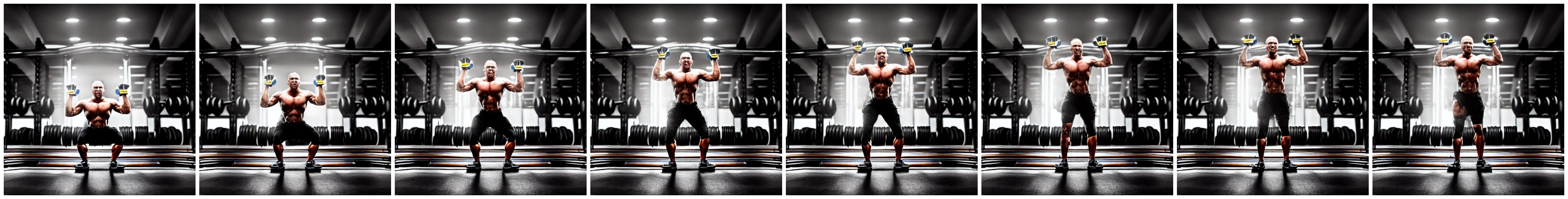} &
        \includegraphics[width=0.32\textwidth, trim={2098 0 1054 0},clip]{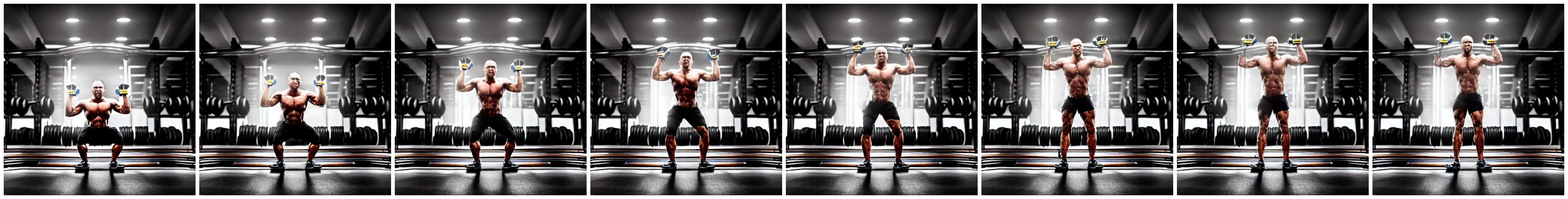} &
        \includegraphics[width=0.32\textwidth, trim={2098 0 1054 0},clip]{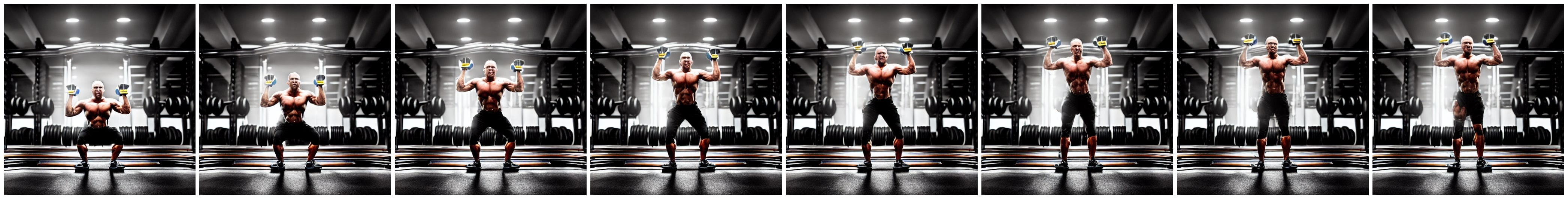} \\
        
        \includegraphics[width=0.32\textwidth, trim={2098 0 1054 0},clip]{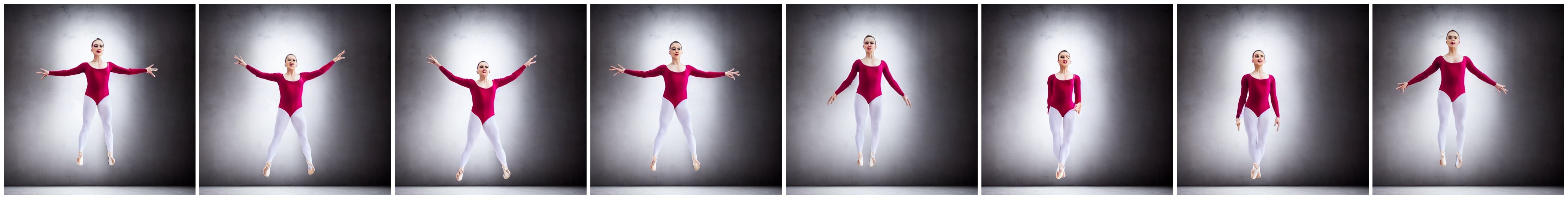} &
        \includegraphics[width=0.32\textwidth, trim={2098 0 1054 0},clip]{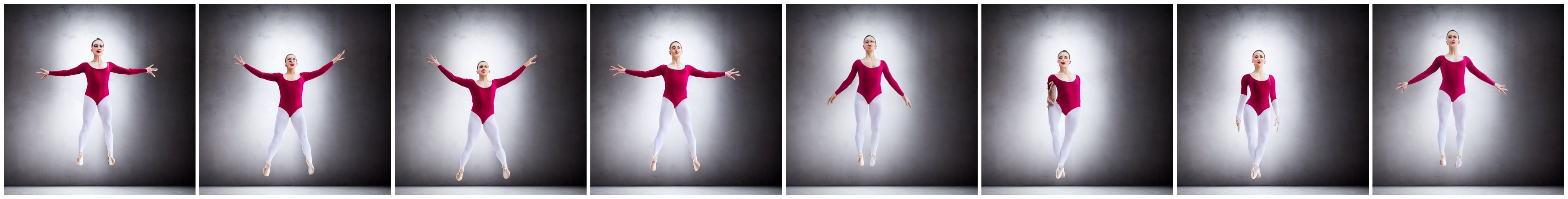} &
        \includegraphics[width=0.32\textwidth, trim={2098 0 1054 0},clip]{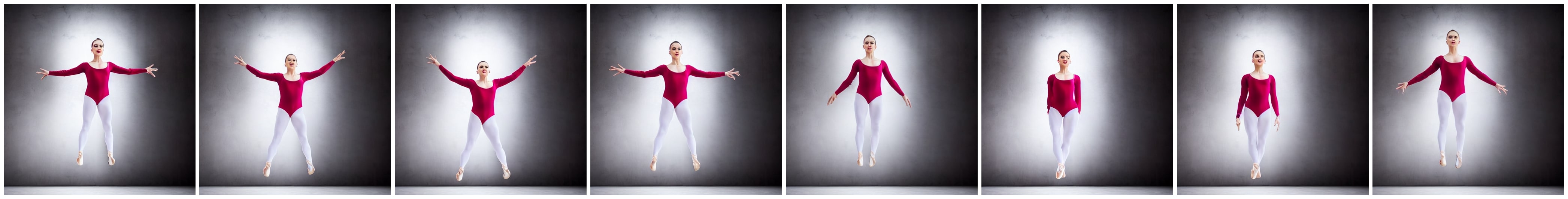} \\
        
        \includegraphics[width=0.32\textwidth, trim={0 0 3152 0},clip]{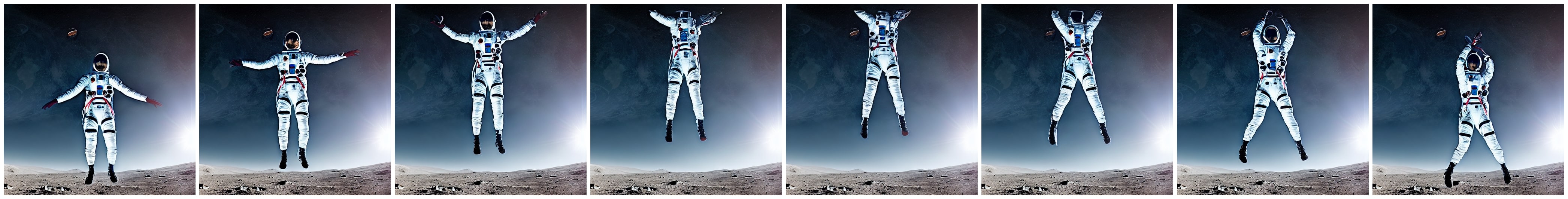} &
        \includegraphics[width=0.32\textwidth, trim={0 0 3152 0},clip]{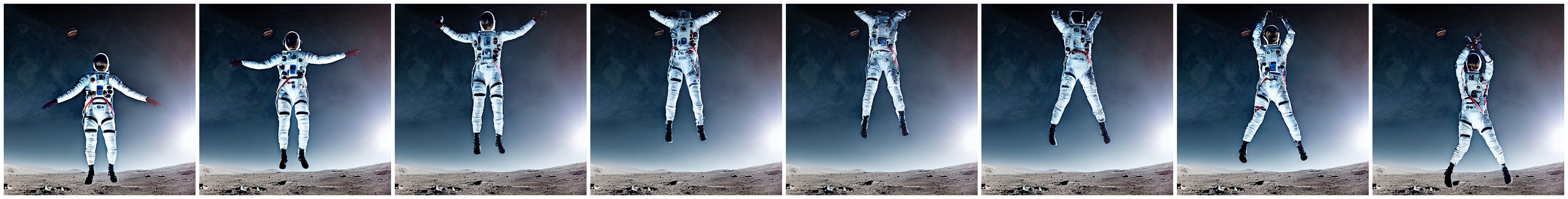} &
        \includegraphics[width=0.32\textwidth, trim={0 0 3152 0},clip]{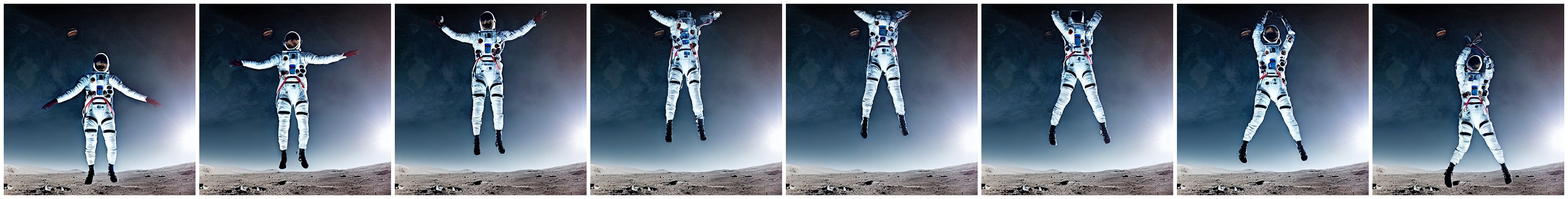} \\
    \end{tabular}
    \caption{An ablation study for different components of our pipeline: Spatial Latent Alignment (SLA), and Pixel-Wise Guidance (PWG).}
    \label{fig:abl}
\end{figure*}

We provide some qualitative examples for the ablation study to complement the quantitative ablation that was provided in the main paper.
\Cref{fig:abl} shows some examples of the contribution of Latent Spatial Alignment (SLA) and Pixel-Wise Guidance (PWG) to achieving temporal consistency.
SLA contributes the most and globally harmonizes the video frames.
However, it can miss some details, especially at the finer parts of the character, \eg the legs in the first row, and the hands in the second and third row.
PWG complements SLA and improves the consistency of the finer parts as it operates on a higher resolution.
It is worth mentioning that employing only PWG is not sufficient to achieve reasonable consistency, but only when combined with SLA, can it impact the output.
The reason is that SLA aligns the latents making the task of PWG easier in improving the fine details.



\setlength{\tabcolsep}{3pt}


\setlength{\tabcolsep}{3pt}

\setlength{\tabcolsep}{3pt}
